%% file: main.tex
\documentclass{article} %
\usepackage{iclr2027_conference,times}

\input{math_commands.tex}

\usepackage{url}

\usepackage{graphicx}
\usepackage{booktabs}
\usepackage{multirow}
\usepackage{wrapfig}
\usepackage{float}
\usepackage{amsmath}
\usepackage{amssymb}
\usepackage{pifont}
\usepackage{etoolbox}
\usepackage{algorithm}
\usepackage{algpseudocode}
\usepackage{hyperref}
\AtBeginEnvironment{table}{\setlength{\abovecaptionskip}{0pt}\setlength{\belowcaptionskip}{5pt}}
\AtEndEnvironment{table}{\vspace{-4pt}}
\AtBeginEnvironment{table*}{\setlength{\abovecaptionskip}{0pt}\setlength{\belowcaptionskip}{5pt}}
\AtEndEnvironment{table*}{\vspace{-4pt}}
\title{Good Rankings, Poor Choices?\\Rethinking Language-Only Vision--Language Model Selection}

\author{Hyuk Kwon, Jaebyeong Jeon, Hyeonseo Jang, Kibok Lee \\
Yonsei University \\
\texttt{\{kh12043, jaebyeong98, jhyeonseo715, kibok\}@yonsei.ac.kr}
}

\newcommand{\cmark}{\ding{51}}
\newcommand{\xmark}{\ding{55}}

\iclrfinalcopy
\begin{document}

\maketitle

\begin{abstract}
Language-Only Vision--Language Model Selection (LOVM) aims to identify the Vision--Language Model (VLM) with the highest zero-shot accuracy on a target task using only textual information.
We revisit the existing LOVM evaluation protocol and argue that ranking metrics, such as top-5 recall and Kendall's rank correlation, do not directly assess the performance of the selected model.
We therefore adopt simple regret (SR), defined as the performance gap between the best and selected models,
and introduce Outlier-Robust Normalized Regret (ORNR) to account for the varying performance ranges across tasks while reducing sensitivity to outliers.
Building on this revised evaluation protocol, we propose Target-Conditioned Expected Source Regret (TC-ESR), a training-free method that selects the model with the lowest similarity-weighted mean regret across source tasks.
Across 22 target tasks and 43 VLMs, TC-ESR achieves the lowest mean SR and ORNR among all compared methods. 
Separately, we identify two sources of target performance leakage in the prior ranking-based evaluation protocol and provide corrections that remove both.
\end{abstract}

\section{Introduction}
\label{sec:intro}

\begin{figure}[!ht]
\centering
\includegraphics[width=0.99\textwidth]{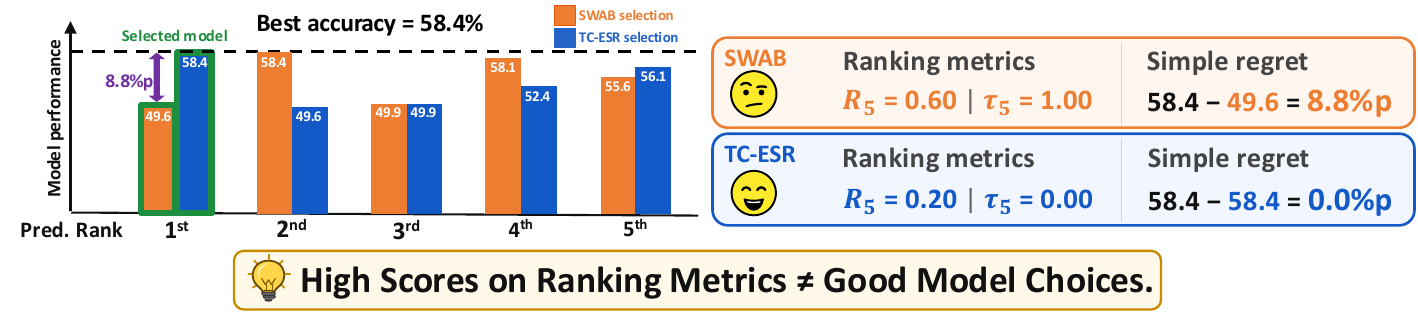}
\caption{Example based on the top-5 rankings produced by SWAB and TC-ESR on GTSRB. SWAB attains higher $R_5$ and $\tau_5$, whereas TC-ESR selects the best-performing model.}
\label{fig:fig1}
\end{figure}

Large-scale image--text pretraining has produced a growing family of vision--language models (VLMs) with strong zero-shot transfer across diverse visual recognition tasks~\citep{radford2021clip,zhai2023siglip}.
As this model zoo expands, selecting an appropriate VLM for each target task becomes increasingly important, since the best model can vary across tasks~\citep{zohar2023lovm}.
Conventional model selection, however, usually requires evaluating the candidates on a labeled dataset for the target task.
Constructing such a dataset entails collecting and annotating target images, which can require substantial time and resources.

Language-Only Vision--Language Model Selection (LOVM) was introduced to avoid this requirement~\citep{zohar2023lovm}.
It assumes no access to target images, labels for individual target instances, or the performance of any candidate model on the target task.
Using textual information of the target task and previously evaluated source tasks, together with candidate model accuracies on the source tasks, an LOVM method predicts a ranking of the candidate models.
Since prior LOVM methods typically induce a ranking over candidate models, they are commonly evaluated by comparing the predicted ranking with the ground-truth ranking determined by target-task accuracy.
In particular, prior work primarily evaluates predicted rankings using top-5 recall and Kendall's rank correlation~\citep{zohar2023lovm,yi2024swab}, which measure top-5 set recovery and relative ordering.
However, these metrics do not directly assess the performance of the selected model on the target task.
As Figure~\ref{fig:fig1} illustrates, higher scores on both ranking metrics need not lead to the selection of a model with higher target accuracy.

To align evaluation with the goal of selecting the best model for the target task from a set of candidate models~\citep{raschka2020modeselection,karimi2021online}, we directly compare the selected model with the best candidate.
Accordingly, we adopt simple regret (SR) to measure the gap between their target accuracies~\citep{bubeck2009regret}.
However, the same SR value can have different implications across tasks because candidate accuracies vary over different ranges.
Normalizing SR by the range of candidate accuracies can make regret values more comparable across tasks, but using the full range is sensitive to extreme candidates: a single unusually weak model can inflate the range and make the normalized value appear artificially small.
We therefore introduce Outlier-Robust Normalized Regret (ORNR), which normalizes SR using a robust performance range computed separately for each task.
ORNR complements SR with a relative measure of the accuracy gap, making regret values more comparable across tasks while reducing the influence of unusually weak candidate models.

This focus on the performance of the selected model also motivates using simple regret as a criterion for model selection.
We first formulate Expected Source Regret (ESR), a simple training-free baseline that selects the model with the lowest mean simple regret across source tasks.
Although ESR attains lower mean regret than prior LOVM methods in our experiments, it gives all source tasks equal weight and therefore does not account for their relevance to the target.
Semantic similarity between class names alone may not fully capture this relevance, since transferability can depend on both domain and task differences~\citep{tan2021otce}.
We therefore extend ESR to Target-Conditioned Expected Source Regret (TC-ESR), which measures source–target similarity along visual domain and label granularity in addition to label semantics, using textual task information alone, and weights source tasks accordingly.
Across 22 target tasks and 43 VLMs, TC-ESR achieves the lowest mean SR and ORNR among all compared methods.

Separately, we identify and correct two sources of target performance leakage in prior evaluation protocols, both arising from the failure to separate the source and target tasks:
First, for each target task, prior methods perform feature selection using all tasks, including the evaluation target. We address this by extending the leave-one-out strategy to feature selection.
Second, ImageNet-1K serves as an evaluation target, even though the ImageNet-1K accuracies of candidate models are used as input features. 
We eliminate this leakage by excluding ImageNet-1K from the target tasks.

We summarize our contributions as follows:
\begin{itemize}
    \item \textbf{Evaluation metrics.}
    We adopt SR to quantify the performance gap between the best and selected models, and introduce ORNR to account for the varying performance ranges across tasks while reducing sensitivity to outliers.

    \item \textbf{Evaluation protocol corrections.}
    We identify and correct two sources of target performance leakage in the prior protocol by performing feature selection exclusively on source tasks and excluding ImageNet-1K from the evaluation targets.
    
    \item \textbf{Selection method.}
    We propose TC-ESR, which uses only textual task information to measure the similarity between source and target tasks along three axes: visual domain, label semantics, and label granularity; it then weights source tasks by these similarities when aggregating the simple regret of each candidate model.

\end{itemize}

\section{Preliminaries}
\label{sec:preliminaries}

\subsection{Problem Setup}
\label{sec:problem_setup}
Given a VLM zoo $\mathcal M$ comprising $M$ candidate models, Language-Only Vision--Language Model Selection (LOVM) aims to identify the best-performing VLM on a target task $t$ using only textual task information, without access to target images or instance-level labels.
The available textual task information consists of a domain and task description $x_t$ and a set of class names $\mathcal C_t$.
In the conventional LOVM setting~\citep{zohar2023lovm,yi2024swab}, candidate VLMs are evaluated on a set of source tasks $\mathcal S_t$ that excludes the target task $t$.
LOVM can then be performed by measuring the similarity between the source tasks $s\in\mathcal S_t$ and the target task $t$, and estimating the relative performance of the candidate VLMs on $t$ by aggregating their performance across $\mathcal S_t$.
For evaluation, a leave-one-out strategy is applied to a set of tasks, with each task serving as $t$ in turn and the remaining tasks as $\mathcal S_t$.

\subsection{Existing LOVM Methods}
\label{sec:existing_lovm}
ModelGPT is the first LOVM method introduced by \citet{zohar2023lovm}.
It constructs features for each candidate VLM using textual proxies for target images, and employs a learnable ranker model to predict the ranking of candidate VLMs on the target task.
SWAB~\citep{yi2024swab} is a subsequent method developed under the same protocol, which uses optimal transport to construct features that capture modality and capability gaps, and further transfers rankings from relevant source classes to rank models.

\section{Rethinking LOVM Evaluation}
\label{sec:rethinking_lovm}

\subsection{Ranking-Based Evaluation}
\label{sec:ranking_eval}
Prior work evaluates LOVM methods using Top-5 Recall ($R_5$) and Kendall's rank correlation ($\tau_5$).
$R_5$ measures the proportion of ground-truth top-5 models correctly identified, whereas $\tau_5$ assesses the relative ordering of models shared between the predicted and ground-truth top-5 rankings.
Although these metrics provide useful diagnostics of LOVM results in terms of ranking quality, they do not directly assess the performance of the selected model on the target task.
In practical model selection scenarios, only the top-ranked model is deployed rather than a ranked set; therefore, these metrics fail to quantify how much the selected model underperforms the best model, as illustrated in Figure~\ref{fig:fig1}.
Table~\ref{tab:metric_disagreement} shows that this discrepancy is systematic across methods and targets.

\subsection{Target Performance Leakage}
\label{sec:target_leakage}
We first examine whether the original LOVM evaluation protocol suffers from information leakage, specifically whether the set of source tasks $\mathcal S_t$ and the target task $t$ are properly separated during evaluation.
First, for each $t$, ModelGPT and SWAB perform feature selection to construct the input features fed to their ranker models based on mean ranking performance over the set of source tasks.
However, this set includes $t$ during feature selection, i.e., $\mathcal S_t \cup \{t\}$.
Consequently, the ground-truth model ranking on $t$ influences feature selection, even though $t$ is intended to be held out when fitting the ranker models.
Second, both methods use the ImageNet-1K accuracies of the candidate models as input features to their ranker models, while ImageNet-1K itself also serves as an evaluation target under the original protocol.
In this case, the input feature reveals the accuracies of the candidate models on $t$, and consequently, their ground-truth ranking.

To eliminate the first leakage, we extend the leave-one-out strategy to feature selection, ensuring that the feature subset for $t$ is selected using only $\mathcal S_t$, without access to $t$.
To eliminate the second leakage, we exclude ImageNet-1K from the target task candidates, while retaining it as a source task and as a dataset for computing input features for ModelGPT and SWAB.
As shown in Table~\ref{tab:eval_protocol_ablation}, both corrections lower $R_5$ and $\tau_5$ of ModelGPT and SWAB, indicating that the original protocol overestimates ranking performance relative to the revised protocol.

\subsection{Regret-Based Evaluation}
\label{sec:regret_eval}
Because model selection aims to identify the candidate with the highest performance on the target task~\citep{raschka2020modeselection,karimi2021online}, we evaluate the selected model by comparing its performance with that of the best candidate.
We use simple regret (SR) to quantify this absolute performance gap~\citep{bubeck2009regret,Renggli_2022_CVPR}.
We further introduce Outlier-Robust Normalized Regret (ORNR) to normalize the gap using a robust performance range defined per task.

\paragraph{Simple Regret.}
Let $m_t^\star=\arg\max_{m\in\mathcal M}a_t(m)$
denote a best-performing model for a task $t$, where $a_t(m)$ is the performance of $m$ on $t$.
We then define the simple regret of selected model $m\in\mathcal M$ on task $t$ as
\begin{equation}
\mathrm{SR}_t(m)
=
a_t(m_t^\star)-a_t(m).
\label{eq:simple-regret}
\end{equation}
Because $a_t(m)\in[0,1]$, $\mathrm{SR}_t(m)\in[0,1]$; zero indicates that a best-performing model is selected, while larger values indicate a greater performance gap.

\paragraph{Outlier-Robust Normalized Regret.}
\begin{figure}[t]
\centering
\includegraphics[width=0.98\columnwidth]{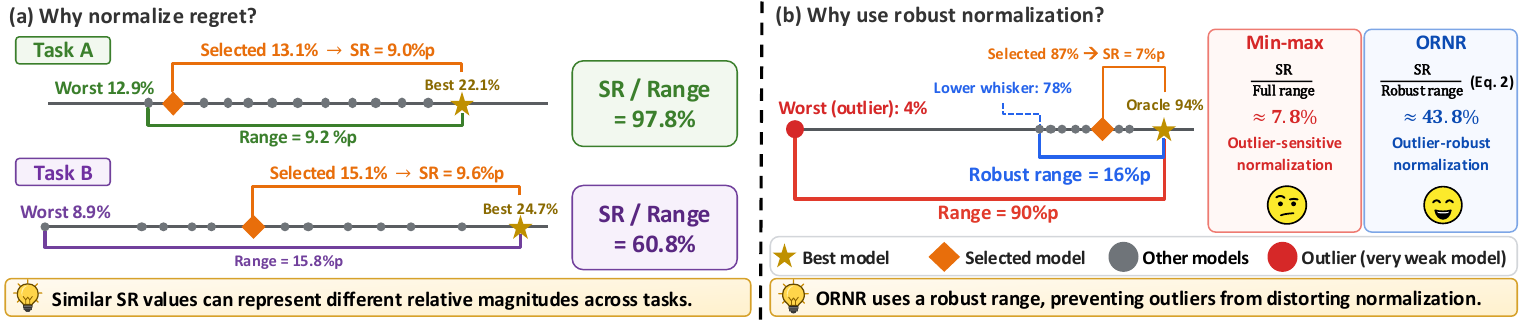} 
\caption{Motivation for ORNR. (a) Similar SR can occupy different fractions of the accuracy range across tasks. (b) An unusually weak model can inflate the full range and distort normalization.}
\label{fig:ornr}
\end{figure}

Although SR directly quantifies the performance loss incurred by the selected model, it does not account for variation in the range of candidate model accuracies across tasks.
As illustrated in Figure~\ref{fig:ornr}(a), similar SR values can have different implications across tasks, representing a small fraction of the accuracy range on one task but a large fraction on another.
Consequently, SR does not provide a common scale for comparing model selection outcomes across tasks.

A straightforward approach is min--max normalization, which divides SR by the range of candidate model accuracies of task $t$:
$\Delta_t^{\mathrm{mm}}
= \max_{u\in\mathcal M}a_t(u) - \min_{u\in\mathcal M}a_t(u).$
However, min--max normalization is sensitive to a single model with unusually low accuracy.
As Figure~\ref{fig:ornr}(b) illustrates, a model with unusually low accuracy can increase the normalization denominator and thereby reduce the normalized value, even though the performance gap between model $m$ and the best model remains unchanged.

We therefore replace the minimum accuracy with the lower whisker of a box-and-whisker plot~\citep{tukey1977eda}.
Let $Q_{1,t}$ and $Q_{3,t}$ denote the first and third quartiles of
$\{a_t(u):u\in\mathcal M\}$.
Following Tukey's boxplot convention, we use the 1.5 IQR rule, a standard nonparametric criterion for flagging potential outliers that requires neither a distributional assumption nor tuning for individual tasks.
We define the interquartile range $\mathrm{IQR}_t=Q_{3,t}-Q_{1,t}$ and the lower fence $F_t^{-}=Q_{1,t}-1.5\,\mathrm{IQR}_t$.
Candidate accuracies below $F_t^{-}$ are treated as potential low outliers.
We then define the lower whisker $\ell_t=\min_{u\in\mathcal M:\,a_t(u)\ge F_t^{-}}a_t(u)$ and the robust performance range $\Delta_t^{\mathrm{rob}}=a_t(m_t^\star)-\ell_t$.
Here, $\ell_t$ is the smallest observed candidate accuracy at or above the lower fence, and $\Delta_t^{\mathrm{rob}}$ is the resulting range between the lower whisker and the accuracy of the best model.
Accordingly, we define ORNR as SR normalized by this robust range:
\begin{equation}
\mathrm{ORNR}_t(m) = \frac{\mathrm{SR}_t(m)}
{\max\!\left(\Delta_t^{\mathrm{rob}},\varepsilon\right)},
\label{eq:robust-regret}
\end{equation}
where $\varepsilon>0$ is a small numerical constant that prevents division by zero.
The lower fence and lower whisker are used only to compute the normalization denominator; they do not remove any model from the candidate set or restrict which model may be selected.
ORNR can exceed one when the selected model's accuracy falls below the lower whisker, because its gap from the best model then exceeds the robust performance range.
We report SR and ORNR together: SR measures the absolute accuracy gap, whereas ORNR scales this gap by the robust range of candidate model accuracies for each task, making regret values more comparable across tasks.
Appendix~\ref{app:robust_ornr} confirms that the comparison between methods is unchanged under alternative denominators
and that ORNR is substantially less sensitive than min--max normalization to weak candidates, although,
like any task-relative normalization, it is not invariant to the composition of the candidate pool.

\section{Method}
\label{sec:method}

\begin{figure}[t]
\centering
\includegraphics[width=0.98\textwidth]{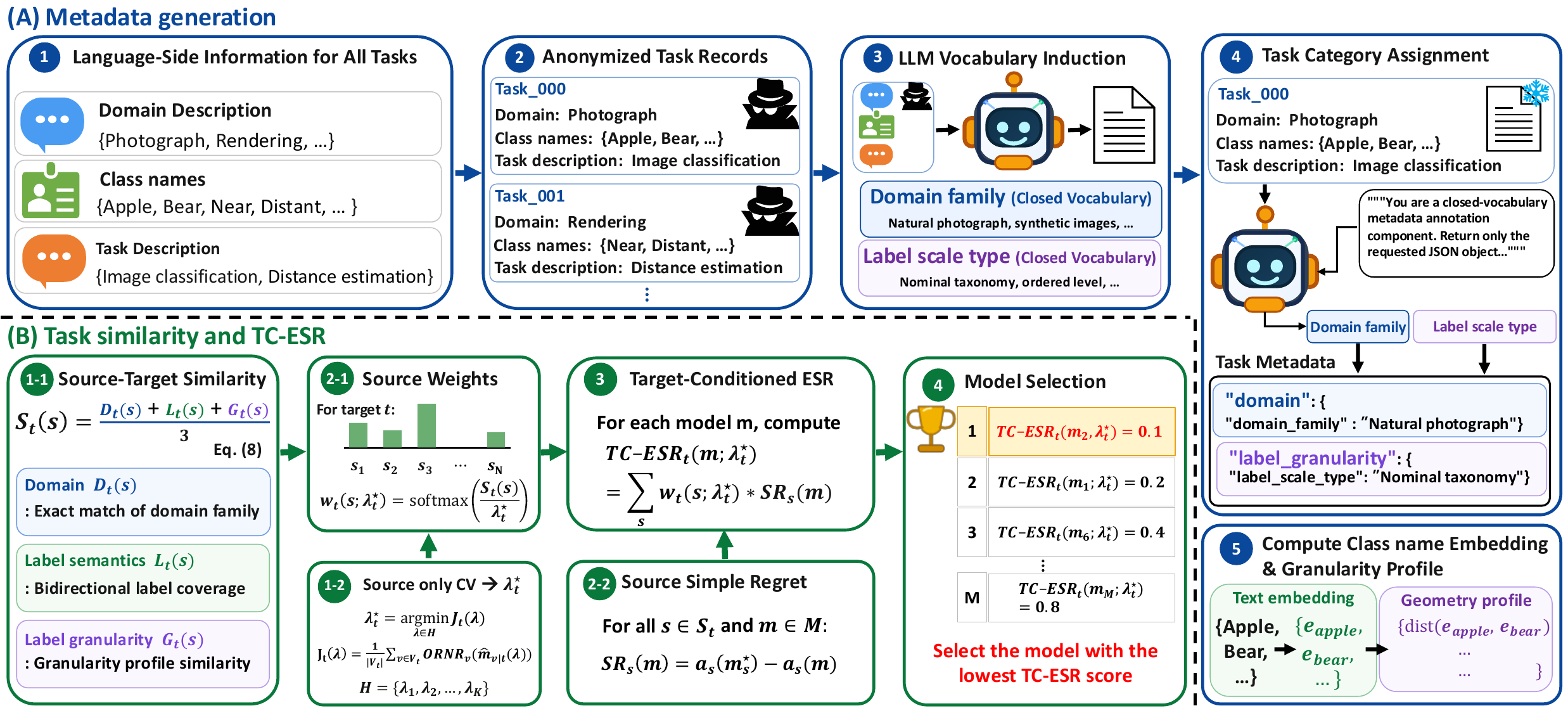} 
\caption{TC-ESR framework.
Text-only task specifications are mapped to three-axis source--target similarities using frozen vocabularies and text embeddings.
$\lambda_t^\star$ is selected by using cross validation over source tasks, and the resulting weights aggregate each model's source task simple regrets. The model with the lowest TC-ESR score is selected.} 
\label{fig:framework}
\end{figure}

Building on this revised evaluation, we use simple regret on source tasks as the basis for model selection. 
\emph{Expected Source Regret} (ESR) selects the model with the lowest mean simple regret across source tasks.
To account for differences in source relevance, we estimate the similarity between each source and the target from text descriptions and class name sets along three axes: visual domain, label semantics, and label granularity.
\emph{Target-Conditioned Expected Source Regret} (TC-ESR) uses these similarities to weight source tasks 
and selects the model with the lowest weighted mean simple regret.
Unlike prior LOVM methods, ESR and TC-ESR do not require training a ranker model.

\subsection{Expected Source Regret}

For each source task $s\in\mathcal{S}_t$ and candidate model $m\in\mathcal{M}$, 
let $\mathrm{SR}_s(m)$ denote the simple regret of selecting $m$ on $s$.
We define the ESR score of model $m$ for target $t$ as its mean simple regret across the available source tasks, and the ESR method selects the candidate with the lowest ESR score:
\begin{equation}
    \mathrm{ESR}_t(m)
    =
    \frac{1}{|\mathcal{S}_t|}
    \sum_{s\in\mathcal{S}_t}
    \mathrm{SR}_s(m),
    \qquad
    \widehat m_t^{\mathrm{ESR}}
    =
    \arg\min_{m\in\mathcal{M}}
    \mathrm{ESR}_t(m).
\end{equation}
A lower ESR score indicates a smaller average gap from the best model on each source task.
Since the best accuracy on each source task is constant across candidate models, this criterion is equivalent to
selecting the model with the highest average source accuracy. 

\subsection{Language-Only Task Similarity}
\label{sec:task_sim}
ESR assigns equal weight to all source tasks, even though tasks more closely related to the target may provide more informative evidence about relative model performance~\citep{zamir2018taskonomy,achille2019task2vec}.
SWAB uses class name similarity to identify relevant source classes when transferring model rankings~\citep{yi2024swab}.
However, semantic similarity between class names alone may not fully capture the relevance of the corresponding source task, since transferability can depend on both domain and task differences~\citep{tan2021otce}.
We therefore estimate the similarity of each source task to the target along three axes informed by prior work: visual domain~\citep{cui2018largescale,tan2021otce}, label semantics~\citep{frome2013devise}, and label granularity~\citep{min2020hierarchical}.
Unlike DCD~\citep{yang2026dcd}, which derives task similarity from unlabeled source and target images, our measure uses only the text descriptions and class name sets of the source and target tasks.

\paragraph{Task Metadata Construction.}

To support the domain and label granularity components of task similarity, we represent each task using two categorical fields: a domain family for visual domain similarity and a label scale type that indicates whether label granularity can be meaningfully estimated from class name geometry.
To ensure that the same set of labels is used for each field across tasks, we use an LLM to induce a shared vocabulary for each field.
This design follows prior work in which LLMs establish a shared label space before assigning labels~\citep{wan2024tntllm,huang2025llmcls}.

Let $\mathcal D=\mathcal S_t\cup\{t\}$ denote the set containing target task $t$ and all source tasks available for it.
In the first stage, the LLM induces a vocabulary for each field, including category labels, definitions, and assignment criteria, from anonymized records $\{(x_d,C_d)\}_{d\in\mathcal D}$, after which we fix both vocabularies.

Second, for each task, the LLM receives only its anonymized record and the two fixed vocabularies and assigns one category from each.
To reduce sensitivity to stochastic outputs, we repeat the category assignment for each task and select the majority category across runs for each field.
For each task $d\in\mathcal D$, let $z_d^{\mathrm{dom}}$ and $z_d^{\mathrm{scale}}$ denote the resulting domain family and label scale assignments, respectively.
We keep these assignments fixed throughout evaluation and compute all similarity scores deterministically from them and the class name embeddings.

\paragraph{Domain Similarity.}
For each source task $s\in\mathcal S_t$, we define its domain similarity to target task $t$ as
\begin{equation}
    D_t(s)
    =
    \mathbb{I}\!\left[
        z_s^{\mathrm{dom}}=z_t^{\mathrm{dom}}
    \right].
\end{equation}
Because the closed vocabulary defines discrete families, we use exact membership rather than graded similarity between family names, as families with semantically similar names need not contain visually similar tasks.

\paragraph{Label Semantic Similarity.}
For each task $d\in\mathcal D$, let $n_d=|C_d|$, and let $c_{d,i}$ denote the class name in $C_d$ indexed by $i$.
Using a fixed text encoder $E$ (e.g., MPNet~\citep{song2020mpnet}), we obtain an $\ell_2$-normalized embedding for each class name:
$\mathbf e_{d,i}=E(c_{d,i})/\lVert E(c_{d,i})\rVert_2\in\mathbb R^p$,
with embedding dimension $p$.

For source task $s\in\mathcal S_t$ and target task $t$, 
we define the pairwise class similarity between $c_{s,i}$ and $c_{t,j}$ as their cosine similarity rescaled to $[0,1]$, $\Gamma^{s,t}_{ij}=(\mathbf e_{s,i}^{\top}\mathbf e_{t,j}+1)/2$.
Because the two class name sets may differ in size or semantic scope, we measure coverage in both directions by matching each class name in one task to its most similar counterpart in the other:
\begin{equation}
\begin{aligned}
\kappa_{s\rightarrow t}
=
\frac{1}{n_t}
\sum_{j=1}^{n_t}
\max_{1\leq i\leq n_s}
\Gamma^{s,t}_{ij},
\
\kappa_{t\rightarrow s}
=
\frac{1}{n_s}
\sum_{i=1}^{n_s}
\max_{1\leq j\leq n_t}
\Gamma^{s,t}_{ij}.
\end{aligned}
\end{equation}
Here, $\kappa_{s\rightarrow t}$ averages the best source match for each target class name, while $\kappa_{t\rightarrow s}$ measures the
reverse.
Their mean defines the label semantic similarity:
\begin{equation}
L_t(s)
=
\frac{\kappa_{s\rightarrow t}+\kappa_{t\rightarrow s}}{2}.
\end{equation}

When one class name set corresponds to only a subset of the other, directional coverage may be high in one direction but low in the reverse.
Averaging the two directions therefore prevents such partial correspondence from producing a high $L_t(s)$.

\paragraph{Label Granularity Similarity.}

Prior work characterizes dataset granularity using distances between classes~\citep{cui2019measuring}.
Since LOVM provides no target images, we estimate granularity from distances between the class name embeddings defined above.
This estimate is intended for taxonomies of semantic categories, rather than labels that denote ordered levels or bins of an underlying attribute.
We therefore set $\eta_d=1$ when $z_d^{\mathrm{scale}}$ indicates such a taxonomy and $\eta_d=0$ otherwise.
For a task $d$ with $\eta_d=1$ and $n_d$ class names, let $\mathcal P_d=\{1-\mathbf e_{d,i}^{\top}\mathbf e_{d,j}:1\le i<j\le n_d\}$ denote the collection of cosine distances between distinct class names.

Let $\widetilde{\mu}_d$ denote the median of $\mathcal P_d$, which provides a robust summary of the pairwise distances between class name embeddings within task $d$.
For a target task with $\eta_t=1$, we define its label granularity similarity to source task $s$ by comparing their median pairwise embedding distances:
\begin{equation}
G_t(s)=
\begin{cases}
\displaystyle
1-
\frac{
\left|\widetilde{\mu}_s-\widetilde{\mu}_t\right|
}{
\max\!\left(
\left|\widetilde{\mu}_s\right|,
\left|\widetilde{\mu}_t\right|,
\varepsilon
\right)
},
& \eta_s=1, \\
0,
& \eta_s=0.
\end{cases}
\end{equation}
Here, $\varepsilon$ is the numerical constant defined above.
Thus, $G_t(s)$ compares the median pairwise embedding distances of the two tasks when the source task also has a taxonomy of semantic categories, and is zero otherwise.

\paragraph{Aggregated Task Similarity.}
We define $S_t(s)$ as a target-conditioned task similarity score for source task $s$ and target task $t$, where larger values indicate greater similarity.
We average the components applicable to the target with equal weights, avoiding additional hyperparameters:
\begin{equation}
    S_t(s)
    =
    \begin{cases}
        \displaystyle
        \frac{D_t(s)+L_t(s)+G_t(s)}{3},
        & \eta_t=1,\\
        \displaystyle
        \frac{D_t(s)+L_t(s)}{2},
        & \eta_t=0.
    \end{cases}
\label{eq:agg_sim}
\end{equation}

Because the included components depend on the target, $S_t(s)$ need not equal $S_s(t)$, consistent with asymmetric transfer relationships~\citep{zamir2018taskonomy,daroya2024task2box}.

\subsection{Target-Conditioned Expected Source Regret}

TC-ESR preserves ESR's modelwise simple regret on source tasks but replaces uniform averaging over source tasks with target-conditioned
weighting.
For a temperature parameter $\lambda>0$, we convert the task similarities into normalized source weights and select the model with the lowest weighted source regret:
\begin{equation}
w_t(s;\lambda)
=
\frac{\exp\left(S_t(s)/\lambda\right)}
{\sum_{u\in\mathcal S_t}\exp\left(S_t(u)/\lambda\right)},
\qquad
\widehat m_t^{\mathrm{TC\text{-}ESR}}
=
\arg\min_{m\in\mathcal M}
\sum_{s\in\mathcal S_t}
w_t(s;\lambda_t^\star)\,\mathrm{SR}_s(m).
\label{eq:tcesr_weight}
\end{equation}
The temperature $\lambda$ controls the concentration of the source weights and is a hyperparameter, which we select for each target from a grid $\mathcal{H}$ by leave-one-out validation over the source tasks.
Consistent with the evaluation targets, ImageNet-1K is used only as a source and not as a validation task, so the validation tasks are $V_t = S_t \setminus \{t_{\mathrm{IN}}\}$, where $t_{\mathrm{IN}}$ denotes ImageNet-1K.
For each $\lambda \in \mathcal{H}$ and each $v \in V_t$, we treat $v$ as the target, compute the weights $w_{v|t}(s;\lambda)$ in Equation~\ref{eq:tcesr_weight} over $S_{t,v} = S_t \setminus \{v\}$, and select $\widehat{m}_{v|t}(\lambda) = \arg\min_{m \in \mathcal{M}} \sum_{s \in S_{t,v}} w_{v|t}(s;\lambda)\,\mathrm{SR}_s(m)$.
We then set $\lambda_t^\star = \arg\min_{\lambda \in \mathcal{H}} J_t(\lambda)$, where $J_t(\lambda) = \frac{1}{|V_t|} \sum_{v \in V_t} \mathrm{ORNR}_v\big(\widehat{m}_{v|t}(\lambda)\big)$ is the mean ORNR.

\section{Experiments}
\label{sec:experiments}
\subsection{Experimental Setup}
\label{sec:setup}

\paragraph{Datasets and Model Zoo.}
Following SWAB~\citep{yi2024swab}, we use the same pool of 23 datasets and 43 candidate VLMs, covering diverse visual recognition tasks and model architectures. 
Complete lists of datasets and models are provided in Appendix~\ref{app:benchmark_details}.

\paragraph{Baselines.}
We compare TC-ESR with the ImageNet baseline (INB) and ModelGPT~\citep{zohar2023lovm}, as well as AvgRank and
SWAB~\citep{yi2024swab}.
INB ranks the candidates by their ImageNet-1K accuracies, whereas AvgRank ranks them by their mean rank across source classes.
We use SWAB's publicly available implementation for all baselines.
ModelGPT and SWAB select their input features by $R_5 + \tau_5$, following their original protocols.

\paragraph{Evaluation Protocol.}
Following prior LOVM work~\citep{zohar2023lovm,yi2024swab}, 
we perform leave-one-out evaluation over the 22 targets excluding ImageNet-1K under our revised evaluation protocol.
For each target, all remaining datasets, including ImageNet-1K, form the source pool.
We evaluate the selected model using SR and ORNR, defined in \eqref{eq:simple-regret} and~\eqref{eq:robust-regret}, respectively, and report macro-averages across targets.
For readability, both metrics are multiplied by 100 and reported in percentage points (\%p) and percent (\%), respectively.

\paragraph{Implementation Details.}
We use \texttt{gpt-5.4-nano} to induce the closed vocabularies and assign a category to each task.
We use \texttt{all-mpnet-base-v2} as the fixed text encoder $E$.
The sensitivity of TC-ESR to the LLM used for metadata construction and to the text encoder is analyzed in Appendix~\ref{app:llm_encoder}.
For each target, we select $\lambda$ from $\mathcal H=\{0.03,0.05,0.07,0.10,0.20,0.30,0.50,0.70\}$ by cross validation over source tasks.
We use $\varepsilon=10^{-12}$ wherever needed to prevent division by zero.
We use the zero-shot accuracies released by SWAB~\citep{yi2024swab} to compute source regrets and evaluate the selected models.
Additional details are provided in Appendix~\ref{app:add_impl_detail}.

\subsection{Main Results}

\begin{table}[t]
\begin{minipage}[t]{0.37\textwidth}
\centering
\caption{Mean SR and ORNR over 22 target tasks.}
\label{tab:main_table}
\setlength{\tabcolsep}{3pt}
\small
\begin{tabular}{@{}lcc@{}}
\toprule
Method
& SR (\%p) $\downarrow$ & ORNR (\%) $\downarrow$ \\
\midrule
INB
& 7.23 & 24.66 \\
AvgRank
& 7.35 & 25.69 \\
ModelGPT
& 7.11 & 24.25 \\
SWAB
& 7.39 & 25.81 \\
\midrule
ESR
& $\underline{5.83}$ & $\underline{22.56}$ \\
TC-ESR
& \textbf{4.66} & \textbf{17.81} \\
\midrule
Best fixed model  & 5.52  & 21.55 \\
\bottomrule
\end{tabular}%
\end{minipage}\hfill
\begin{minipage}[t]{0.55\textwidth}
\centering
\caption{Ablation of the three task similarity axes used in TC-ESR.}
\label{tab:component_analysis}
\setlength{\tabcolsep}{3pt}
\small\renewcommand{\arraystretch}{1.065}\vspace*{1.7pt}
\begin{tabular}{@{}cccc|cc@{}}
\toprule
\multicolumn{3}{c}{Task Similarity Axes} & \multicolumn{2}{c}{Performance} \\
\cmidrule(lr){1-3} \cmidrule(lr){4-5}
$D_t(s)$ & $L_t(s)$ & $G_t(s)$ & SR (\%p) $\downarrow$ & ORNR (\%) $\downarrow$ \\
\midrule
\cmark & \cmark & \cmark & \textbf{4.66} & \textbf{17.81} \\
\midrule
\xmark & \cmark & \cmark & 5.91 & 23.05 \\
\cmark & \xmark & \cmark & 7.39 & 26.39 \\
\cmark & \cmark & \xmark & 8.22 & 28.58 \\
\bottomrule
\end{tabular}%
\end{minipage}
\end{table}

As shown in Table~\ref{tab:main_table}, ESR attains lower mean SR and ORNR than all prior methods, suggesting that simple regret on source tasks provides more useful evidence for model selection than source ranks.
TC-ESR, which differs from ESR only in its source weights, further reduces both metrics,
indicating that emphasizing source tasks more similar to the target improves model selection over uniform aggregation.
TC-ESR and ESR remain the best and second-best methods when ModelGPT and SWAB instead select their input features by SR or ORNR, as shown in Appendix~\ref{app:baseline_retune}.
TC-ESR also attains lower mean SR and ORNR than the best fixed model, the single candidate with the highest mean target accuracy across targets, although this reference requires target accuracies unavailable in LOVM.
Appendix~\ref{app:discussion} further analyzes where target conditioning improves model selection. 

\subsection{Ablation Studies}

\paragraph{Analysis of Task Similarity Components.}
\label{sec:analysis}

Table~\ref{tab:component_analysis} evaluates the three axes of the task similarity measure over 22 target tasks, with label granularity applied to the 13 targets for which it is applicable.
Removing any component degrades both metrics, with the largest degradation observed when label granularity is removed.
Using all three axes performs best, indicating that the three axes make complementary contributions within the TC-ESR procedure.

\paragraph{Effect of Source Weight Assignment.}
\begin{table}[t]
\begin{minipage}[t]{0.44\textwidth}
\centering
\caption{Effect of source weight assignment. Permuted: averaged over 10K permutations.}
\label{tab:weight_assignment}
\setlength{\tabcolsep}{3pt}
\small
\begin{tabular}{@{}lcc@{}}
\toprule
Weight Assignment & SR (\%p) $\downarrow$ & ORNR (\%) $\downarrow$ \\
\midrule
Uniform (ESR)
& 5.83 & 22.56 \\
Similarity (TC-ESR)
& \textbf{4.66} & \textbf{17.81} \\
Permuted
& 9.08 & 30.64 \\
\bottomrule
\end{tabular}%
\end{minipage}\hfill
\begin{minipage}[t]{0.52\textwidth}
\centering
\caption{Comparison with alternative uses of the same task information.}
\label{tab:simple_baselines}
\setlength{\tabcolsep}{3pt}
\small
\begin{tabular}{@{}lcc@{}}
\toprule
Method & SR (\%p) $\downarrow$ & ORNR (\%) $\downarrow$ \\
\midrule
Uniform (ESR)          & 5.83 & 22.56 \\
Nearest source         & 6.43 & 24.75 \\
Top-$k$ sources        & 9.31 & 29.56 \\
Description embedding  & 5.95 & 23.25 \\
Class names only       & 6.52 & 23.26 \\
Learned weights        & 5.36 & 22.70 \\
\midrule
TC-ESR                 & \textbf{4.66} & \textbf{17.81} \\
\bottomrule
\end{tabular}%
\end{minipage}
\end{table}

To test whether TC-ESR benefits from unequal weights themselves or from assigning larger weights to more similar sources, we randomly permute each target's weights across source tasks, keeping the set of weight values unchanged.
As shown in Table~\ref{tab:weight_assignment}, similarity-based assignment outperforms uniform weighting, whereas random assignment falls below even uniform weighting.
Thus, the improvement comes from assigning larger weights to more similar sources, supporting the premise of TC-ESR that models performing well on source tasks similar to the target also tend to perform well on the target.

\paragraph{Comparison with Alternative Uses of Task Information.}

We next compare TC-ESR with alternative uses of the same task information (Table~\ref{tab:simple_baselines}; details in Appendix~\ref{app:add_impl_detail}).
These alternatives use the TC-ESR similarity to select only a few sources (\emph{Nearest source}, \emph{Top-$k$ sources}), replace it with a single task embedding (\emph{Description embedding}) or with label semantic similarity alone (\emph{Class names only}), or learn the weights of the three axes by source-only validation instead of averaging them (\emph{Learned weights}).
None of them outperforms ESR on both metrics.
Restricting aggregation to the most similar sources performs worse than weighting all sources, suggesting that less similar sources also provide useful evidence, and the simpler similarities do not improve over uniform weighting.
Together with Table~\ref{tab:weight_assignment}, these results show that the gains of TC-ESR depend on how the similarity is defined and applied across all source tasks, not merely on similarity-based weighting.

\subsection{Further Analysis}

\paragraph{Ranking Metrics vs. Selection Quality.}
\begin{table}[t]
\centering
\caption{Agreement between ranking metrics and SR over six methods and 22 targets.}
\label{tab:metric_disagreement}
\small
\begin{tabular}{lcccc}
\toprule
Ranking metric & Rank corr.\ with SR & Disagreement (\%) & Mean SR loss (\%p) & Median SR loss (\%p) \\
\midrule
$R_5$    & $-0.83$ & 25.7 & 3.32 & 1.82 \\
$\tau_5$ & $-0.47$ & 43.1 & 2.46 & 1.05 \\
\bottomrule
\end{tabular}
\end{table}

Table~\ref{tab:metric_disagreement} examines whether the discrepancy illustrated in Figure~\ref{fig:fig1} is systematic and how much target accuracy it costs.
First, we rank the six LOVM methods in Table~\ref{tab:main_table} by their mean ranking metric and by their mean SR, negating SR so that higher is better for both, and compute the Kendall rank correlation between the two rankings.
For $R_5$, the correlation is $-0.83$, so among these six methods, those with higher mean $R_5$ selected worse models on average.
Second, for each target and each pair of methods, excluding ties, we count a \emph{disagreement} when the method with the higher ranking score has the higher SR, and define the \emph{SR loss} as the SR gap between the two methods.
$R_5$ and $\tau_5$ disagree with SR in 25.7\% and 43.1\% of the comparisons, with mean SR losses of 3.32\%p and 2.46\%p, so choosing between methods by their ranking scores often favors the one that selects the worse model, and $\tau_5$ disagrees nearly as often as a random choice would.
This negative relationship is consistent with how the methods are designed: prior methods are built to predict rankings and select their input features by $R_5 + \tau_5$, whereas ESR and TC-ESR directly minimize the regret of the selected model on source tasks.
Retuning the prior methods by SR does not reverse this relationship, as shown in Appendix~\ref{app:baseline_retune}.
These results support evaluating LOVM methods by the regret of the selected model rather than by ranking metrics, and suggest that selection methods should target this regret directly.

\paragraph{Effect of the Protocol Corrections.}

Although SR and ORNR are our primary metrics for evaluating the final model choice, $R_5$ and $\tau_5$ remain useful for diagnosing how the protocol corrections affect the rankings predicted by prior methods.
In Table~\ref{tab:eval_protocol_ablation}, ``Feature selection'' denotes source only feature selection, whereas ``Exclude IN-1K'' removes ImageNet-1K from the target set.
Avg. change denotes the mean relative change from the original protocol, averaged over $R_5$ and $\tau_5$ for both ModelGPT and SWAB.
Each correction lowers both metrics for both methods, and applying both produces the largest decrease in all four cases.

\begin{table}[t]
\caption{Effect of protocol corrections on $R_5$ and $\tau_5$ for ModelGPT and SWAB.}
\label{tab:eval_protocol_ablation}
\centering
\small
\begin{tabular}{@{}lccccc@{}}
\toprule
\multirow{2}{*}{Leakage removed} & \multicolumn{2}{c}{ModelGPT} & \multicolumn{2}{c}{SWAB} & \multirow{2}{*}{Avg. change} \\
\cmidrule(lr){2-3}
\cmidrule(lr){4-5}
& $R_5$ $\uparrow$ & $\tau_5$ $\uparrow$ & $R_5$ $\uparrow$ & $\tau_5$ $\uparrow$ & \\
\midrule
None (original) & 0.443 & 0.267 & 0.504 & 0.320 & -- \\
Feature selection & 0.435 & 0.252 & 0.496 & 0.302 & -3.7\% \\
Exclude IN-1K & 0.418 & 0.233 & 0.482 & 0.294 & -7.7\% \\
Both (final) & 0.411 & 0.180 & 0.474 & 0.283 & -14.3\% \\
\bottomrule
\end{tabular}
\end{table}

\section{Conclusion}

We revisited LOVM evaluation and showed that ranking metrics often disagree with the performance of the selected model, which SR and ORNR measure directly, and corrected two sources of target performance leakage in the prior protocol.
Under this revised protocol, our proposed TC-ESR achieves the lowest mean SR and ORNR by weighting the simple regret on source tasks with their similarity to the target, measured solely from text along three axes.
These results suggest that LOVM methods should be evaluated and designed around the performance of the selected model rather than the quality of the predicted ranking.

\input{main.bbl}
\appendix
\section{Related Work}

\subsection{Model Selection and LOVM}

Conventional model selection evaluates candidate models on a labeled target validation set~\citep{raschka2020modeselection}.
Transferability estimators such as LEEP and LogME instead use labeled target examples together with candidate predictions or representations to estimate transfer performance without adapting every model~\citep{nguyen2020leep,you2021logme}.

\citet{Renggli_2022_CVPR} formulate pretrained model search under a limited fine tuning budget and evaluate search methods using regret, but their proposed hybrid method uses labeled downstream data.
These approaches therefore do not address the setting in which no target images are available.

Language-Only Vision--Language Model Selection (LOVM) selects a pretrained VLM using textual target task information and performance observed on source tasks~\citep{zohar2023lovm}.
ModelGPT~\citep{zohar2023lovm} constructs textual proxies for the target classes and learns a ranker model from evaluated source tasks.
SWAB~\citep{yi2024swab} uses optimal transport to transfer source class statistics and rankings while accounting for modality and capability gaps.
Prior LOVM work evaluates the predicted ranking mainly with Top-5 Recall and Kendall's $\tau_5$.
We instead measure the performance gap between the selected model and the best candidate, and use task similarity derived from text to weight simple regret on the source tasks.

\subsection{Other VLM Selection Settings}

MLL~\citep{tan2025mll} addresses a related setting but differs from LOVM in the information it uses and how it selects models.
It first constructs a semantic graph with commonly occurring visual concepts and images from the sample datasets, and performs model labeling by evaluating each candidate VLM on this graph.
For a target task, MLL uses these labels to select a model separately for each class and combines the selected models' predictions during inference.
It may therefore use different candidate VLMs for different classes, whereas LOVM selects one candidate VLM for the entire target task.
Thus, the output of MLL does not correspond to the single selected model evaluated in the LOVM setting.

DCD~\citep{yang2026dcd} directly compares with prior LOVM methods, but considers a different information setting in which unlabeled images from both the source and target tasks are available.
It processes these images with each candidate visual encoder to construct model specific task representations and estimate directional source relevance.
LOVM, by contrast, assumes that no target images are available and relies on textual information about the target task.

\citet{li2026rethinkingmodelselectionvlm} 
select a vision encoder for a fixed LLM before multimodal training and estimate the performance of the resulting VLM.
This differs from selecting one complete pretrained VLM for a downstream task without additional training.

\subsection{Task Similarity for Source Weighting}

Task relationships have been estimated using trained task models and transfer experiments, labeled examples and probe networks, or source and target data~\citep{zamir2018taskonomy,achille2019task2vec,tan2021otce}.
Task2Box further models asymmetric task relationships with box embeddings~\citep{daroya2024task2box}.
However, these approaches require images, labels, probe models, or observed transfer outcomes.

Prior work suggests that visual domain proximity, label semantics, and label granularity are important factors in transferability~\citep{cui2018largescale,tan2021otce,frome2013devise,cui2019measuring,min2020hierarchical}.

Motivated by these findings, TC-ESR estimates task similarity along these three axes using only task and domain descriptions and class name sets.
The resulting target conditioned similarity is shared across candidate VLMs and weights their simple regrets on source tasks, without target images or candidate model inference.

\section{LOVM Benchmark and Evaluation Details}
\label{app:benchmark_details}

\subsection{Datasets}
\label{app:dataset_details}

Following prior LOVM work~\citep{zohar2023lovm,yi2024swab}, 
we use a collection $\mathcal D$ of 23 tasks covering a diverse range of visual recognition tasks and image domains.
Table~\ref{tab:dataset_details} reports the number of classes and the task and domain descriptions released with the LOVM benchmark.
Following prior LOVM studies~\citep{zohar2023lovm,yi2024swab}, we assume access to these descriptions and the class name sets.
We use this information to construct the task metadata for TC-ESR.
Under our revised evaluation protocol, ImageNet-1K is retained in the source pool but is not used as an evaluation target.
Specifically, letting $t_{\mathrm{IN}}$ denote ImageNet-1K, the evaluation target set is
$\mathcal T=\mathcal D\setminus\{t_{\mathrm{IN}}\}$.
For each target $t\in\mathcal T$, all remaining tasks form the source pool,
$\mathcal S_t=\mathcal D\setminus\{t\}$.
Thus, evaluation is performed over 22 target tasks, while ImageNet-1K remains available as a source task for every target.

\begin{table*}[t]
\centering
\caption{The task information follows the LOVM and SWAB benchmarks~\citep{zohar2023lovm,yi2024swab}.}
\label{tab:dataset_details}
{%
\fontsize{9}{10}\selectfont
\setlength{\tabcolsep}{8pt}
\renewcommand{\arraystretch}{1.02}
\begin{tabular}{@{}cccc@{}}
\toprule
Dataset & Classes & Task & Domain \\
\midrule
Imagenet~\citep{deng2009imagenet} & 1000 & classification  & natural image \\
SUN397~\citep{xiao2010sun} & 397  & scene und.      & natural image \\
Country211~\citep{radford2021clip} & 211  & geolocation     & natural image \\
Stanford Cars~\citep{krause2013car} & 196  & classification  & natural image \\
Flowers102~\citep{nilsback2008flower} & 102  & classification  & natural image \\
CIFAR100~\citep{krizhevsky2009learning} & 100  & classification  & natural image \\
DTD~\citep{cimpoi2014dtd} & 46   & classification  & texture image \\
RESISC45~\citep{cheng2017remotesense} & 45   & classification  & satellite images \\
GTSRB~\citep{stallkamp2011gtsrb}    & 43   & classification  & natural image \\
Oxford Pets~\citep{parkhi2012pets}   & 37   & classification  & natural image \\
VOC2007~\citep{everingham2007voc}       & 20   & classification  & natural image \\
STL10~\citep{coates2011stl10}          & 10   & classification  & natural image \\
EuroSAT~\citep{helber2019satellites}       & 10   & classification  & satellite images \\
MNIST~\citep{lecun2010mnist}         & 10   & classification  & hand-writing \\
SVHN~\citep{Netzer2011svhn}          & 10   & OCR             & natural image \\
CLEVR-C~\citep{johnson2017clevr}       & 8    & object counting & natural image \\
CLEVR-D~\citep{johnson2017clevr}       & 8    & distance est.   & natural image \\
FER2013~\citep{goodfellow2013challengesrepresentationlearningreport}       & 7    & fac. exp. rec.  & natural image \\
DMLab~\citep{zhai2020dmlab}         & 6    & distance est.   & synthetic \\
Retinopathy~\citep{kaggle2015retinopathy}   & 5    & classification  & retina scan \\
KITTI~\citep{geiger2013kitti}         & 4    & distance est.   & natural image \\
PCam~\citep{veeling2018pcam}          & 2    & classification  & histopathology \\
Rendered SST2~\citep{radford2021clip} & 2    & OCR             & text image \\
\bottomrule
\end{tabular}
}
\end{table*}

\subsection{VLM Zoo}
\label{sec:vlm_zoo_details}

We adopt the 43 candidate VLMs evaluated by SWAB~\citep{yi2024swab}, comprising OpenCLIP, BLIP, and BEiT-3 checkpoints, as summarized in Table~\ref{tab:model_zoo}.
We use the zero-shot accuracies released by SWAB for all 43 models and 23 datasets.
These accuracies are used as source task performance by the LOVM methods and as target task performance only for final evaluation, so we do not rerun zero-shot inference.

\begin{table*}[!t]
\centering
\caption{Detailed information on the 43 candidate models used in the LOVM benchmark.
The model information follows the LOVM and SWAB benchmarks~\citep{zohar2023lovm,yi2024swab}.}
\label{tab:model_zoo}
\begingroup
\setlength{\tabcolsep}{4pt}
\renewcommand{\arraystretch}{0.95}
\resizebox{\textwidth}{!}{%
\begin{tabular}{@{}r|cc|cc@{}}
\toprule
ID & Model & Name & Dataset & Name \\
\midrule
1  & RN50                         & RN50                         & openai                                      & WIT \\
2  & RN101                        & RN101                        & openai                                      & WIT \\
3  & RN50x4                       & RN50x4                       & openai                                      & WIT \\
4  & RN50x16                      & RN50x16                      & openai                                      & WIT \\
5  & RN50x64                      & RN50x64                      & openai                                      & WIT \\
6  & ViT-B-32                     & ViT-B/32                     & laion400m\_e31                              & L400m \\
7  & ViT-B-32                     & ViT-B/32                     & laion400m\_e32                              & L400m \\
8  & ViT-B-32-quickgelu           & ViT-B/32                     & laion400m\_e32                              & L400m \\
9  & ViT-B-32                     & ViT-B/32                     & openai                                      & WIT \\
10 & ViT-B-32                     & ViT-B/32                     & laion2b\_s34b\_b79k                         & L2b-b \\
11 & ViT-B-32                     & ViT-B/32                     & laion2b\_e16                                & L2b-c \\
12 & ViT-B-16                     & ViT-B/16                     & laion400m\_e32                              & L400m \\
13 & ViT-B-16                     & ViT-B/16                     & openai                                      & WIT \\
14 & ViT-B-16-240                 & ViT-B/16-240                 & laion400m\_e32                              & L400m \\
15 & ViT-L-14                     & ViT-L/14                     & laion400m\_e31                              & L400m \\
16 & ViT-L-14                     & ViT-L/14                     & laion400m\_e32                              & L400m \\
17 & ViT-L-14                     & ViT-L/14                     & laion2b\_s32b\_b82k                         & L2b-b \\
18 & ViT-L-14                     & ViT-L/14                     & openai                                      & WIT \\
19 & ViT-L-14-336                 & ViT-L/14-336                 & openai                                      & WIT \\
20 & ViT-G-14                     & ViT-G/14                     & laion2b\_s12b\_b42k                         & L2b-a \\
21 & ViT-G-14                     & ViT-G/14                     & laion2b\_s34b\_b88k                         & L2b-a \\
22 & ViT-H-14                     & ViT-H/14                     & laion2b\_s32b\_b79k                         & L2b-b \\
23 & coca\_ViT-B-32               & CoCa-ViT-B/32                & laion2b\_s13b\_b90k                         & L2b-c \\
24 & coca\_ViT-B-32               & CoCa-ViT-B/32                & mscoco\_finetuned\_laion2b\_s13b\_b90k     & L2b-c + coco \\
25 & coca\_ViT-L-14               & CoCa-ViT-L/14                & laion2b\_s13b\_b90k                         & L2b-c \\
26 & coca\_ViT-L-14               & CoCa-ViT-L/14                & mscoco\_finetuned\_laion2b\_s13b\_b90k     & L2b-c + coco \\
27 & convnext\_base               & ConvNEXT-B                   & laion400m\_s13b\_b51k                       & L400m-c \\
28 & convnext\_base\_w            & ConvNEXT-BW                  & laion2b\_s13b\_b82k                         & L2b-d \\
29 & convnext\_base\_w            & ConvNEXT-BW                  & laion2b\_s13b\_b82k\_augreg                 & L2b-e \\
30 & convnext\_base\_w            & ConvNEXT-BW                  & laion\_aesthetic\_s13b\_b82k                & L2b-f \\
31 & convnext\_base\_w\_320       & ConvNEXT-BW-320              & laion\_aesthetic\_s13b\_b82k                & L2b-f \\
32 & convnext\_base\_w\_320       & ConvNEXT-BW-320              & laion\_aesthetic\_s13b\_b82k\_augreg        & L2b-g \\
33 & convnext\_large\_d           & ConvNEXT-LD                  & laion2b\_s26b\_b102k\_augreg                & L2b-h \\
34 & convnext\_large\_d\_320      & ConvNEXT-LD-320              & laion2b\_s29b\_b131k\_ft                    & L2b-i \\
35 & convnext\_large\_d\_320      & ConvNEXT-LD-320              & laion2b\_s29b\_b131k\_ft\_soup              & L2b-j \\
36 & BLIP\_retrieval\_base\_coco  & BLIP\_retrieval\_base\_coco  & COCO+VG+CC+CC12M+SBU                        & BLIP-Dataset \\
37 & BLIP\_retrieval\_base\_f30k  & BLIP\_retrieval\_base\_f30k  & COCO+VG+CC+CC12M+SBU+Flickr30k              & BLIP-Dataset-f \\
38 & BLIP\_retrieval\_large\_coco & BLIP\_retrieval\_large\_coco & COCO+VG+CC+CC12M+SBU                        & BLIP-Dataset \\
39 & BLIP\_retrieval\_large\_f30k & BLIP\_retrieval\_large\_f30k & COCO+VG+CC+CC12M+SBU+Flickr30k              & BLIP-Dataset-f \\
40 & BEiT-3\_retrieval\_base\_coco
   & BEiT-3\_retrieval\_base\_coco
   & CC12M+CC3M+SBU+COCO+VG
   & BEiT-Dataset \\
41 & BEiT-3\_retrieval\_base\_f30k
   & BEiT-3\_retrieval\_base\_f30k
   & CC12M+CC3M+SBU+COCO+VG+Flickr30k
   & BEiT-Dataset-f \\
42 & BEiT-3\_retrieval\_large\_coco
   & BEiT-3\_retrieval\_large\_coco
   & CC12M+CC3M+SBU+COCO+VG
   & BEiT-Dataset \\
43 & BEiT-3\_retrieval\_large\_f30k
   & BEiT-3\_retrieval\_large\_f30k
   & CC12M+CC3M+SBU+COCO+VG+Flickr30k
   & BEiT-Dataset-f \\
\bottomrule
\end{tabular}%
}
\endgroup
\end{table*}

\subsection{Ranking Metrics}
\label{sec:ranking_metric_details}

In addition to SR and ORNR, we report the ranking metrics used in prior LOVM evaluation: Top-5 Recall and Kendall's rank correlation~\citep{zohar2023lovm,yi2024swab}.
These metrics measure agreement between the predicted and ground truth model rankings.
They are retained as supplementary diagnostics of ranking quality, whereas SR and ORNR are our primary metrics for evaluating the model selected at rank one.

For target task $t$, let $\widehat{\mathbf r}_t$ denote the predicted ranking of the candidate models and let $\mathbf r_t^\star$ denote the ground truth ranking obtained by sorting the models by decreasing zero-shot accuracy on $t$.

For both rankings, rank one denotes the most highly ranked model.
For methods defined through a score to be minimized, including ESR and TC-ESR, the predicted ranking is obtained by sorting candidate models in increasing order of their scores.
For any ranking $\mathbf r$, define its top-5 model set as
\begin{equation}
\mathcal T_5(\mathbf r)
=
\left\{
m\in\mathcal M : r(m)\leq 5
\right\}.
\end{equation}
The models shared by the predicted and ground truth top-5 sets are
\begin{equation}
\mathcal I_t
=
\mathcal T_5(\widehat{\mathbf r}_t)
\cap
\mathcal T_5(\mathbf r_t^\star).
\end{equation}

\paragraph{Top-5 Recall.}
Top-5 recall measures the proportion of the ground truth top-5 models recovered in the predicted top-5 set:
\begin{equation}
R_5(t)
=
\frac{|\mathcal I_t|}{5}.
\end{equation}
Thus, $R_5(t)\in[0,1]$, with a value of one indicating that the predicted and ground truth top-5 sets contain the same models.
The metric evaluates set recovery and does not depend on the ordering of the recovered models.

\paragraph{Kendall's Rank Correlation.}
Kendall's rank correlation measures agreement between the relative orderings of the models shared by the two top-5 sets.
Following the released LOVM implementation, it is computed only over $\mathcal I_t$:
\begin{equation}
\tau_5(t)
=
\begin{cases}
\displaystyle
\tau\!\left(
\widehat{\mathbf r}_t|_{\mathcal I_t},
\mathbf r_t^\star|_{\mathcal I_t}
\right),
&
|\mathcal I_t|\geq 3, \\[4pt]
0,
&
|\mathcal I_t|<3,
\end{cases}
\end{equation}
where $\tau(\cdot,\cdot)$ denotes Kendall's rank correlation and
$\mathbf r|_{\mathcal I_t}$ denotes the relative ordering restricted to the models in $\mathcal I_t$.
The metric ranges from $-1$ to $1$, where larger values indicate stronger agreement.
We use the notation $\tau_5$ to make explicit that the correlation is computed only among models appearing in both top-5 sets.
For both metrics, we report the macro mean across the 22 evaluation targets:
\begin{equation}
\overline{R}_5
=
\frac{1}{|\mathcal T|}
\sum_{t\in\mathcal T}R_5(t),
\qquad
\overline{\tau}_5
=
\frac{1}{|\mathcal T|}
\sum_{t\in\mathcal T}\tau_5(t).
\end{equation}
Higher values are better for both metrics.

\subsection{Revised Evaluation Protocol}

\begin{algorithm}[t]
\caption{Nested source-only feature selection for
ModelGPT and SWAB under the revised LOVM protocol.}
\label{alg:revised_protocol}

\begin{algorithmic}[1]
\Require Task collection $\mathcal D$; ImageNet task
$t_{\mathrm{IN}}$; target
$t\in\mathcal D\setminus\{t_{\mathrm{IN}}\}$
\Require Candidate feature configurations $\mathcal F$;
validation metric $Q$, where larger is better
\Ensure Predicted ranking $\hat r_t$

\State $\mathcal T \gets \mathcal D \setminus \{t_{\mathrm{IN}}\}$
\State $\mathcal S_t \gets \mathcal D \setminus \{t\}$
\State $\mathcal V_t \gets \mathcal T \setminus \{t\}$

\For{each configuration $f \in \mathcal F$}
    \State $q_f \gets 0$
    \For{each validation task $v \in \mathcal V_t$}
        \State $\mathcal S_{t,v}
        \gets \mathcal D \setminus \{t,v\}$
        \State $\hat r_v^{\,f,\mathcal S_{t,v}}
        \gets
        \Call{FitRankerAndPredict}
        {f,\mathcal S_{t,v},v}$
        \State $q_f \gets q_f
        + Q\!\left(
        \hat r_v^{\,f,\mathcal S_{t,v}},
        r_v^\star
        \right)$
    \EndFor
    \State $q_f \gets q_f / |\mathcal V_t|$
\EndFor

\State $\hat f_t
\gets \operatorname*{arg\,max}_{f\in\mathcal F} q_f$
\State $\hat r_t
\gets
\Call{FitRankerAndPredict}{\hat f_t,\mathcal S_t,t}$

\State \Return $\hat r_t$
\end{algorithmic}
\end{algorithm}

Algorithm~\ref{alg:revised_protocol} summarizes the revised evaluation protocol for ModelGPT and SWAB, which fit a ranker model using only source task data.
For ModelGPT, the candidate feature configurations are
$\mathcal F_{\mathrm{ModelGPT}}=\{\mathrm{INB}, C, G\}$,
where INB denotes the ImageNet baseline and $C$ and $G$ denote the features introduced in LOVM.
For SWAB, the candidate feature configurations are
$\mathcal F_{\mathrm{SWAB}}
=\{\mathrm{INB}, \mathrm{GAP}_C, \mathrm{GAP}_G\}$,
where $\mathrm{GAP}_C$ and $\mathrm{GAP}_G$ are obtained by incorporating the modality gap estimated by SWAB into $C$ and $G$, respectively.
Following SWAB, we use $Q=R_5+\tau_5$ as the validation metric.
For each target $t$, each candidate configuration is evaluated by treating every $v\in\mathcal V_t$ as a validation task and fitting the ranker using
$\mathcal S_{t,v}=\mathcal D\setminus\{t,v\}$.
The configuration with the highest mean $Q$ over the validation tasks is then used to fit the final ranker model on
$\mathcal S_t=\mathcal D\setminus\{t\}$
and predict the ranking for $t$.
Thus, candidate performance on $t$ is not used for either configuration selection or ranker model fitting.

TC-ESR does not fit a ranker model and is training-free.
To follow the same source-only principle, we apply this nested validation procedure only when selecting its softmax temperature $\lambda$.
For each target $t$, $\lambda$ is selected using the validation tasks in $\mathcal V_t$ and the corresponding source sets
$\mathcal S_{t,v}$, without using candidate performance on $t$.
Since
$\mathcal T=\mathcal D\setminus\{t_{\mathrm{IN}}\}$
and
$\mathcal V_t=\mathcal T\setminus\{t\}$,
ImageNet-1K is used neither as a target nor as a validation task.
It remains available as a source task, including for the input features of ModelGPT and SWAB and the source regrets used by ESR and TC-ESR.

\section{Additional Implementation Details}
\label{app:add_impl_detail}
\paragraph{Computational Resources.} 
All experiments are conducted on a server equipped with an AMD Ryzen 9 7950X CPU with 16 cores and a single NVIDIA
GeForce RTX 4090 GPU with 24 GB of memory. We use PyTorch 2.10.0 with CUDA 12.8.

\paragraph{Baseline Implementation and Random Seeds.}
We obtain the results for INB, AvgRank, ModelGPT~\citep{zohar2023lovm}, and SWAB~\citep{yi2024swab} using the implementation released with SWAB under our revised evaluation protocol.
In the released implementation, ModelGPT introduces stochasticity by adding seed dependent Gaussian noise to the generated text embeddings used to construct its proxy features.
SWAB inherits this stochastic proxy feature construction in its ModelGPT based ranker branch, while its optimal transport rank transfer and final rank aggregation are deterministic.
Because the proxy feature construction produces variation across seeds, we repeat the full evaluation of ModelGPT and SWAB using random seeds $1$ through $10$ and report the mean over the 10 runs.
In contrast, INB and AvgRank rank models directly from fixed performance statistics, while ESR and TC-ESR
deterministically aggregate fixed source regrets once the task metadata and text embeddings are fixed.
We therefore evaluate INB, AvgRank, ESR, and TC-ESR once with the random seed fixed to $1$.

\paragraph{Details of the Alternative Selection Methods.}

All alternatives in Table~\ref{tab:simple_baselines} use the same task metadata, text encoder, and
class-name template as TC-ESR, and select their hyperparameters by the same source-only validation
with mean ORNR as the objective.
Top-$k$ sources ranks source tasks by the TC-ESR similarity $S_t(s)$ and averages SR uniformly over
the top $k$ sources with $k \in \{1, 3, 5, 7\}$, averaging over tied sources; Nearest source
corresponds to $k = 1$.
Description embedding represents each task by the normalized sum of the embedding of its domain and
task description and the mean embedding of its class names, and uses their cosine similarity
rescaled to $[0, 1]$ in place of $S_t(s)$.
Class names only uses $L_t(s)$ in place of $S_t(s)$.
Learned weights replaces the equal average in Equation~\ref{eq:agg_sim} with
$w_D D_t(s) + w_L L_t(s) + w_G G_t(s)$, where the weights are selected jointly with $\lambda$ from a
simplex grid with step 0.25 augmented with equal weights; for targets without the granularity
component, the weights of the remaining two axes are renormalized.
Description embedding, Class names only, and Learned weights use the softmax weighting in
Equation~\ref{eq:tcesr_weight} with $\lambda$ selected from the same grid as TC-ESR.

\section{Task Metadata Construction Details}

TC-ESR represents each task using two categorical fields:
a domain family for visual domain similarity and a label scale type that indicates whether label granularity can be meaningfully estimated from class name geometry. 
We construct the task metadata in two steps: vocabulary induction and task category assignment. 
For each field, the shared vocabulary consists of category labels, definitions, and assignment criteria.

\subsection{Vocabulary Induction}
For vocabulary induction, each anonymized task record contains a domain description, a task description, and all class names. 
The domain and task descriptions for all datasets are reported in Table~\ref{tab:dataset_details}.
Dataset names, images, candidate model accuracies, model rankings, and downstream results are not provided. From the 23 task records, the LLM induces seven categories for the domain family and six categories for the label scale type, together with their definitions and assignment criteria. 
We then fix both vocabularies before task category assignment.
The resulting category labels are listed in Table~\ref{tab:metadata_vocabularies}.

\begin{table*}[t]
\centering
\caption{Category labels in the shared vocabularies for the two task metadata fields.}
\label{tab:metadata_vocabularies}
{%
\setlength{\tabcolsep}{5pt}
\begin{tabular}{@{}lp{0.76\textwidth}@{}}
\toprule
Field & Category labels \\
\midrule
Domain family
&
Object centered natural photographs;
scene centered natural photographs;
texture centered photographs;
document or glyph images;
medical images;
aerial or satellite images;
synthetic or rendered images
\\
Label scale type
&
Unordered semantic taxonomy;
binary state;
ordered level;
numeric count;
symbol or code identity;
multi factor composition
\\
\bottomrule
\end{tabular}
}
\end{table*}

\subsection{Task Category Assignment}
For each task and each run, the LLM receives only its anonymized task record and the two fixed vocabularies and assigns one category from each vocabulary.
To reduce sensitivity to stochastic variation in the LLM outputs, we repeat the category assignment five times per task and select the majority category across runs for each field.
All 23 tasks yield a majority within five runs.
For 21 tasks, all five runs produce the same domain family assignment; for 20 tasks, all five runs produce the same label scale assignment.
Table~\ref{tab:task_metadata} reports the resulting task metadata and agreement across runs.

\begin{table*}[t]
\centering
\caption{Task metadata for the 23 tasks.
Agreement denotes the number of the five runs that assign the selected category.
We set $\eta_d=1$ only when the label scale type indicates a taxonomy of semantic categories and $n_d \geq 3$, and set $\eta_d=0$ otherwise.
The number of classes for each task is reported in Table~\ref{tab:dataset_details}.
ImageNet-1K is included during metadata construction and as a source task, but is excluded from the evaluation targets.}
\label{tab:task_metadata}
{%
\scriptsize
\setlength{\tabcolsep}{3.5pt}
\begin{tabular}{@{}llclccc@{}}
\toprule
Task
& Domain family
& Agreement
& Label scale type
& Agreement
& $\eta_d$ \\
\midrule
Stanford Cars
& Object centered natural photographs
& 5/5
& Unordered semantic taxonomy
& 5/5
& 1 \\

CIFAR100
& Object centered natural photographs
& 5/5
& Unordered semantic taxonomy
& 5/5
& 1 \\

CLEVR closest distance
& Synthetic or rendered images
& 5/5
& Ordered level
& 5/5
& 0 \\

CLEVR count all
& Synthetic or rendered images
& 5/5
& Numeric count
& 5/5
& 0 \\

Country211
& Scene centered natural photographs
& 5/5
& Unordered semantic taxonomy
& 5/5
& 1 \\

Diabetic Retinopathy
& Medical images
& 5/5
& Ordered level
& 5/5
& 0 \\

DMLab
& Synthetic or rendered images
& 5/5
& Multi factor composition
& 3/5
& 0 \\

DTD
& Texture centered photographs
& 5/5
& Unordered semantic taxonomy
& 5/5
& 1 \\

EuroSAT
& Aerial or satellite images
& 5/5
& Unordered semantic taxonomy
& 5/5
& 1 \\

FER2013
& Object centered natural photographs
& 5/5
& Unordered semantic taxonomy
& 5/5
& 1 \\

Flowers102
& Object centered natural photographs
& 5/5
& Unordered semantic taxonomy
& 5/5
& 1 \\

GTSRB
& Scene centered natural photographs
& 4/5
& Unordered semantic taxonomy
& 5/5
& 1 \\

ImageNet-1K
& Object centered natural photographs
& 5/5
& Unordered semantic taxonomy
& 5/5
& 1 \\

KITTI closest distance
& Object centered natural photographs
& 5/5
& Ordered level
& 3/5
& 0 \\

MNIST
& Document or glyph images
& 5/5
& Symbol or code identity
& 5/5
& 0 \\

PCam
& Medical images
& 5/5
& Unordered semantic taxonomy
& 3/5
& 0 \\

Oxford Pets
& Object centered natural photographs
& 5/5
& Unordered semantic taxonomy
& 5/5
& 1 \\

Rendered SST2
& Document or glyph images
& 5/5
& Binary state
& 5/5
& 0 \\

RESISC45
& Aerial or satellite images
& 4/5
& Unordered semantic taxonomy
& 5/5
& 1 \\

STL10
& Object centered natural photographs
& 5/5
& Unordered semantic taxonomy
& 5/5
& 1 \\

SUN397
& Scene centered natural photographs
& 5/5
& Unordered semantic taxonomy
& 5/5
& 1 \\

SVHN
& Document or glyph images
& 5/5
& Symbol or code identity
& 5/5
& 0 \\

VOC2007
& Object centered natural photographs
& 5/5
& Unordered semantic taxonomy
& 5/5
& 1 \\
\bottomrule
\end{tabular}
}
\end{table*}

\subsection{Class Name Embeddings and Label Granularity}
For all 23 tasks, we encode each class name using \texttt{all-mpnet-base-v2} with the fixed template \texttt{Class label: \{class name\}} and $\ell_2$ normalize the resulting embeddings.
These embeddings are used to compute label semantic similarity for every pair of source and target tasks.
For each task with $\eta_d=1$, we additionally compute
$\widetilde{\mu}_d$ as the median pairwise cosine distance between distinct class name embeddings, following the definition in the main paper. 
Label granularity is therefore computed for 14 of the 23 tasks, including ImageNet-1K, and for 13 of the 22 evaluation targets.

The two vocabularies, the resulting task metadata, and the class name embeddings are generated once and held fixed throughout evaluation. 
The domain family assignment is used to compute domain similarity, while the label scale assignment determines whether the label granularity
component is applicable. 
Label semantic similarity is computed directly from these class name embeddings, independently of the categorical metadata.

\section{More Experimental Results}

\subsection{Further Ablation Studies}

All design choices of TC-ESR were fixed before evaluation on the target tasks, following standard defaults and the motivations given in Section~\ref{sec:task_sim}: exact matching for categorical domain families, the median as a robust summary of pairwise distances, bidirectional coverage to avoid favoring partial correspondence, and an unweighted average of the three axes. The ablations below are post-hoc analyses of these choices rather than the basis on which they were selected.

\begin{table}[t]
\begin{minipage}[t]{0.29\textwidth}
\centering
\caption{Comparison of domain similarity measures.}
\label{tab:domain_similarity_ablation}
\footnotesize
\setlength{\tabcolsep}{2pt}
\begin{tabular}{@{}lcc@{}}
\toprule
& SR & ORNR \\
Measure & (\%p) $\downarrow$ & (\%) $\downarrow$ \\
\midrule
Semantic sim. & 6.71 & 24.43 \\
Exact (Ours) & \textbf{4.66} & \textbf{17.81} \\
\bottomrule
\end{tabular}
\end{minipage}\hfill
\begin{minipage}[t]{0.34\textwidth}
\centering
\caption{Comparison of label semantic aggregation rules.}
\label{tab:label_semantic_aggregation}
\footnotesize
\setlength{\tabcolsep}{2pt}
\begin{tabular}{@{}lcc@{}}
\toprule
& SR & ORNR \\
Aggregation rule & (\%p) $\downarrow$ & (\%) $\downarrow$ \\
\midrule
Mean (all pairs) & 4.83 & 19.59 \\
Balanced OT & 5.14 & 21.41 \\
$\kappa_{s \rightarrow t}$ only & 5.57 & 22.80 \\
$\kappa_{t \rightarrow s}$ only & 5.29 & 21.56 \\
Bidirectional (Ours) & \textbf{4.66} & \textbf{17.81} \\
\bottomrule
\end{tabular}
\end{minipage}\hfill
\begin{minipage}[t]{0.34\textwidth}
\centering
\caption{Comparison of label granularity statistics.}
\label{tab:label_granularity_statistic}
\footnotesize
\setlength{\tabcolsep}{2pt}
\begin{tabular}{@{}lcc@{}}
\toprule
& SR & ORNR \\
Statistic & (\%p) $\downarrow$ & (\%) $\downarrow$ \\
\midrule
NN mean ($k$=1) & 8.21 & 27.45 \\
NN mean ($k$=3) & 8.51 & 29.68 \\
NN mean ($k$=6) & 8.50 & 29.65 \\
Mean dist. & 7.74 & 24.38 \\
Median dist. (Ours) & \textbf{4.66} & \textbf{17.81} \\
\bottomrule
\end{tabular}
\end{minipage}
\end{table}

\begin{table}[t]
\centering
\caption{Mean results over 22 target tasks under the revised LOVM evaluation protocol.
SR is reported in percentage points, whereas ORNR, $R_5$, and $\tau_5$ are reported as percentages.
For ModelGPT and SWAB, each target result is averaged over ten seeds before macro averaging.
Bold and underline denote the best and second-best mean results, respectively.}
\label{tab:main_table_full}
{%
\fontsize{9.0}{10.0}\selectfont
\setlength{\tabcolsep}{5pt}
\begin{tabular}{lcccc}
\toprule
Method 
& SR (\%p) $\downarrow$ & ORNR (\%) $\downarrow$ & $R_5$ (\%) $\uparrow$ & $\tau_5$ (\%) $\uparrow$  \\
\midrule
INB
& 7.23 & 24.66 & \underline{41.8} & \underline{23.3} \\
AvgRank
& 7.35 & 25.69 & \underline{41.8} & 23.0  \\
ModelGPT
& 7.11 & 24.25 & 41.1 & 18.0 \\
SWAB
&  7.39 & 25.81 & \textbf{47.4} & \textbf{28.3} \\
\midrule
ESR
& $\underline{5.83}$ & $\underline{22.56}$ & 31.8 & 18.2 \\
TC-ESR
& \textbf{4.66} & \textbf{17.81} & 40.0 & 22.7 \\
\bottomrule
\end{tabular}
}
\end{table}

\paragraph{Domain Similarity Measure.}
We encode the assigned domain family names using the fixed text encoder $E$. We then compare exact matching of the assigned domain families with cosine similarity between the resulting embeddings.
As shown in Table~\ref{tab:domain_similarity_ablation}, exact matching yields lower SR and ORNR, consistent with the motivation in Section~\ref{sec:task_sim}.
This may be because domain families with semantically similar labels do not necessarily contain visually similar tasks.

\paragraph{Label Semantic Similarity Measure.}

We compare five aggregation rules for deriving $L_t(s)$ from the pairwise class similarity matrix $\Gamma^{s,t}$, while keeping all other components fixed.
The mean over all pairs averages every entry of $\Gamma^{s,t}$, whereas balanced optimal transport couples the two class name sets under uniform marginals~\citep{peyre2019computational}.
The score $\kappa_{s \rightarrow t}$ averages the strongest source match for each target class, while $\kappa_{t \rightarrow s}$ applies the same operation in the reverse direction.
Bidirectional coverage averages these two directional scores.
As shown in Table~\ref{tab:label_semantic_aggregation}, bidirectional coverage achieves the lowest mean SR and ORNR.
These results suggest that bidirectional coverage benefits from retaining each class's strongest match in both directions and evaluating these matches independently.

\paragraph{Label Granularity Measure.}

As shown in Table~\ref{tab:label_granularity_statistic}, the median pairwise distance achieves the lowest SR and ORNR.
The mean over all pairwise distances also outperforms the nearest neighbor statistics, suggesting that overall class separation is more informative than local neighborhood structure.
The further improvement from the mean to the median is consistent with the benefit of reduced sensitivity to extreme pairwise distances.
We test $k\in\{1,3,6\}$; among the tasks for which label granularity is computed, the smallest label set contains seven classes, making $k=6$ the largest candidate.

\paragraph{Regret Metrics for Source Aggregation.}

\begin{table}[t]
\centering
\caption{Effect of the regret used for source aggregation.}
\label{tab:regret_objective_ablation}
\begin{tabular}{@{}lccc@{}}
\toprule
Method
& Source Regret & SR (\%p) $\downarrow$ & ORNR (\%) $\downarrow$ \\
\midrule
\multirow{2}{*}{ESR}
& SR   & 5.83 & 22.56 \\
& ORNR & 7.03 & 27.19 \\
\midrule
\multirow{2}{*}{TC-ESR}
& SR   & \textbf{4.66} & \textbf{17.81} \\
& ORNR & 4.70 & 18.97 \\
\bottomrule
\end{tabular}
\end{table}

Table~\ref{tab:regret_objective_ablation} compares the regret used for source aggregation and for selecting $\lambda$.
For source aggregation, ORNR normalizes each source regret by the performance range of the corresponding task.
This normalization alters source contributions independently of target similarity.
In contrast, SR preserves absolute performance gaps and lets the similarity weights control source contributions.

\paragraph{Sensitivity to the LLM and Text Encoder.}
\label{app:llm_encoder}

\begin{table}[t]
\centering
\begin{minipage}[t]{0.45\linewidth}
\centering
\caption{Sensitivity to the LLM used for metadata construction.}
\label{tab:llm_ablation}
\small
\begin{tabular}{lcc}
\toprule
LLM & SR (\%p) & ORNR (\%) \\
\midrule
gpt-5.4-nano      & 4.66 & 17.81 \\
Qwen3-4B-Instruct & 4.83 & 19.59 \\
\bottomrule
\end{tabular}
\end{minipage}
\hfill
\begin{minipage}[t]{0.5\linewidth}
\centering
\caption{Sensitivity to the text encoder.}
\label{tab:encoder_ablation}
\small
\begin{tabular}{lcc}
\toprule
Text encoder & SR (\%p) & ORNR (\%) \\
\midrule
all-mpnet-base-v2    & 4.66 & 17.81 \\
BGE-base-en-v1.5     & 4.83 & 19.59 \\
all-distilroberta-v1 & 4.88 & 19.79 \\
\bottomrule
\end{tabular}
\end{minipage}
\end{table}

Tables~\ref{tab:llm_ablation} and~\ref{tab:encoder_ablation} report TC-ESR with a different LLM for
metadata construction and with different text encoders, keeping all other components fixed.
Replacing gpt-5.4-nano with the open-source Qwen3-4B-Instruct, we rerun the entire metadata
construction, including vocabulary induction and category assignment. The resulting SR of 4.83\%p
remains lower than those of ESR and the best prior method, ModelGPT, which are 5.83\%p and 7.11\%p,
respectively.
For the text encoder, we evaluate BGE-base-en-v1.5 and all-distilroberta-v1 with encoder-specific
templates. Both encoders yield SR and ORNR comparable to MPNet.

\subsection{Per-Task Experimental Results}
\begin{table*}[t]
\centering
\caption{Per-task results under the revised LOVM evaluation protocol.
SR is reported in percentage points, whereas ORNR, $R_5$, and $\tau_5$ are reported as percentages.
For ModelGPT and SWAB, each per-task result is averaged over ten seeds.
Bold and underline denote the best and second-best means in the Avg. column, respectively.}
\label{tab:per_task_results}
{%
\setlength{\tabcolsep}{2.2pt}
\renewcommand{\arraystretch}{1.05}
\resizebox{\textwidth}{!}{%
\begin{tabular}{@{}ll|*{22}{c}|c@{}}
\toprule
& Method
& \rotatebox{90}{cars}
& \rotatebox{90}{cifar100}
& \rotatebox{90}{clevr-distance}
& \rotatebox{90}{clevr-count}
& \rotatebox{90}{country211}
& \rotatebox{90}{retinopathy}
& \rotatebox{90}{dmlab}
& \rotatebox{90}{dtd}
& \rotatebox{90}{eurosat}
& \rotatebox{90}{fer2013}
& \rotatebox{90}{flowers}
& \rotatebox{90}{gtsrb}
& \rotatebox{90}{kitti}
& \rotatebox{90}{mnist}
& \rotatebox{90}{pcam}
& \rotatebox{90}{pets}
& \rotatebox{90}{renderedsst2}
& \rotatebox{90}{resisc45}
& \rotatebox{90}{stl10}
& \rotatebox{90}{sun397}
& \rotatebox{90}{svhn}
& \rotatebox{90}{voc2007}
& Avg. \\
\midrule

\multirow{6}{*}{SR ($\%$p) $\downarrow$}
& INB
& 0.00 & 0.10 & 5.37 & 11.68 & 3.56 & 49.81 & 9.06
& 0.00 & 5.87 & 2.70 & 0.49 & 8.73 & 12.52 & 13.03
& 9.09 & 0.41 & 14.39 & 0.00 & 0.49 & 1.27 & 10.56
& 0.00 & 7.23 \\

& AvgRank
& 0.46 & 0.00 & 9.59 & 12.85 & 4.61 & 42.93 & 7.75
& 1.22 & 0.00 & 8.16 & 0.00 & 0.00 & 32.49 & 12.47
& 4.62 & 0.00 & 7.08 & 0.57 & 0.96 & 0.09 & 12.38
& 3.37 & 7.35 \\

& ModelGPT
& 0.00 & 0.10 & 5.79 & 11.68 & 3.56 & 49.81
& 8.93 & 0.00 & 5.87 & 3.24 & 0.49 & 6.11
& 12.52 & 13.03 & 8.64 & 0.37 & 13.64 & 0.00
& 0.53 & 1.27 & 10.56 & 0.34 & 7.11 \\

& SWAB
& 0.00 & 0.02 & 9.59 & 11.68 & 3.56 & 49.13
& 9.06 & 0.00 & 5.87 & 2.70 & 0.49 & 8.73
& 12.52 & 13.03 & 8.64 & 0.41 & 10.29 & 0.00
& 0.49 & 1.27 & 12.38 & 2.69 & 7.39 \\

& ESR
& 0.29 & 1.99 & 3.27 & 9.58 & 7.93 & 20.57 & 5.60
& 0.21 & 0.04 & 2.68 & 3.17 & 0.17 & 32.35 & 10.12
& 10.46 & 0.79 & 15.32 & 0.95 & 0.86 & 1.82 & 0.00
& 0.13 & \underline{5.83} \\

& TC-ESR
& 0.00 & 0.10 & 3.27 & 15.16 & 4.61 & 2.22 & 5.60 & 0.00 & 5.87 & 2.70 & 0.49 & 0.00 & 12.52 & 10.12 & 5.94 & 0.41 & 14.94 & 0.57 & 0.49 & 0.09 & 17.52 & 0.00 & \textbf{4.66} \\

\midrule
\multirow{6}{*}{ORNR ($\%$) $\downarrow$}
& INB
& 0.00 & 0.28 & 33.44 & 38.30 & 12.18 & 70.37 & 98.52
& 0.00 & 14.32 & 15.85 & 2.16 & 33.82 & 37.71 & 26.97
& 51.46 & 4.48 & 67.88 & 0.00 & 9.31 & 7.27 & 18.20
& 0.00 & 24.66 \\

& AvgRank
& 1.16 & 0.00 & 59.78 & 42.14 & 15.76 & 60.65 & 84.22
& 4.45 & 0.00 & 48.01 & 0.00 & 0.00 & 97.88 & 25.81
& 26.17 & 0.00 & 33.42 & 2.06 & 18.38 & 0.53 & 21.34
& 23.52 & 25.69 \\

& ModelGPT
& 0.00 & 0.28 & 36.08 & 38.30 & 12.18 & 70.37 & 97.09
& 0.00 & 14.32 & 19.07 & 2.16 & 23.68 & 37.71 & 26.97
& 48.93 & 4.03 & 64.35 & 0.00 & 10.21 & 7.27 & 18.20
& 2.35 & 24.25 \\

& SWAB
& 0.00 & 0.06 & 59.78 & 38.30 & 12.18 & 69.40 & 98.52
& 0.00 & 14.32 & 15.85 & 2.16 & 33.82 & 37.71 & 26.97
& 48.93 & 4.48 & 48.55 & 0.00 & 9.31 & 7.27 & 21.34
& 18.81 & 25.81 \\

& ESR
& 0.72 & 5.48 & 20.40 & 31.41 & 27.14 & 29.06 & 60.88
& 0.77 & 0.09 & 15.73 & 14.05 & 0.68 & 97.46 & 20.94
& 59.27 & 8.66 & 72.28 & 3.43 & 16.47 & 10.42 & 0.00
& 0.93 & \underline{22.56} \\

& TC-ESR
& 0.00 & 0.28 & 20.40 & 49.70 & 15.76 & 3.13 & 60.88 & 0.00 & 14.32 & 15.85 & 2.16 & 0.00 & 37.71 & 20.94 & 33.62 & 4.48 & 70.47 & 2.06 & 9.31 & 0.53 & 30.21 & 0.00 & \textbf{17.81} \\

\midrule

\multirow{6}{*}{$R_5 \ (\%) \uparrow$}
& INB
& 80.0 & 80.0 & 0.0 & 40.0 & 40.0 & 0.0 & 0.0
& 80.0 & 60.0 & 40.0 & 60.0 & 60.0 & 0.0 & 20.0
& 0.0 & 60.0 & 0.0 & 100.0 & 20.0 & 60.0 & 60.0
& 60.0 & \underline{41.8} \\

& AvgRank
& 80.0 & 80.0 & 0.0 & 40.0 & 40.0 & 0.0 & 0.0
& 80.0 & 60.0 & 40.0 & 60.0 & 60.0 & 0.0 & 20.0
& 0.0 & 60.0 & 0.0 & 100.0 & 20.0 & 60.0 & 60.0
& 60.0 & \underline{41.8} \\

& ModelGPT
& 80.0 & 80.0 & 0.0 & 38.0 & 50.0 & 0.0 & 0.0 & 80.0 & 58.0 & 36.0 & 66.0 & 58.0 & 0.0 & 12.0 & 0.0 & 66.0 & 8.0 & 90.0 & 20.0 & 60.0 & 42.0 & 60.0 & 41.1 \\

& SWAB
& 80.0 & 80.0 & 0.0 & 60.0 & 40.0 & 4.0 & 0.0 & 80.0 & 60.0 & 56.0 & 80.0 & 60.0 & 0.0 & 40.0 & 0.0 & 80.0 & 22.0 & 100.0 & 20.0 & 60.0 & 60.0 & 60.0 & \textbf{47.4} \\

& ESR
& 60.0 & 40.0 & 0.0 & 60.0 & 20.0 & 0.0 & 0.0
& 80.0 & 60.0 & 20.0 & 40.0 & 40.0 & 0.0 & 40.0
& 0.0 & 40.0 & 0.0 & 60.0 & 20.0 & 20.0 & 60.0
& 40.0 & 31.8 \\

& TC-ESR
& 80.0 & 60.0 & 20.0 & 0.0 & 20.0 & 20.0 & 0.0 & 80.0 & 60.0 & 40.0 & 60.0 & 20.0 & 0.0 & 40.0 & 20.0 & 60.0 & 40.0 & 60.0 & 40.0 & 60.0 & 40.0 & 60.0 & 40.0 \\

\midrule

\multirow{6}{*}{$\tau_5 \ (\%) \uparrow$}
& INB
& 66.7 & 66.7 & 0.0 & 0.0 & 0.0 & 0.0 & 0.0
& 66.7 & 100.0 & 0.0 & 33.3 & 100.0 & 0.0 & 0.0
& 0.0 & -33.3 & 0.0 & 80.0 & 0.0 & -100.0 & 33.3
& 100.0 & \underline{23.3} \\

& AvgRank
& 33.3 & 100.0 & 0.0 & 0.0 & 0.0 & 0.0 & 0.0
& 33.3 & 100.0 & 0.0 & 100.0 & 100.0 & 0.0 & 0.0
& 0.0 & 33.3 & 0.0 & 40.0 & 0.0 & -33.3 & -33.3
& 33.3 & 23.0 \\

& ModelGPT
& 66.7 & 66.7 & 0.0 & 0.0 & -16.7 & 0.0 & 0.0 & 66.7 & 76.7 & 0.0 & 43.3 & 76.7 & 0.0 & 0.0 & 0.0 & -20.0 & 0.0 & 72.7 & 0.0 & -100.0 & -3.3 & 66.7 & 18.0 \\

& SWAB
& 66.7 & 98.3 & 0.0 & 33.3 & 0.0 & 0.0 & 0.0 & 39.8 & 100.0 & 26.7 & 33.3 & 100.0 & 0.0 & 0.0 & 0.0 & 33.3 & 0.0 & 57.0 & 0.0 & -81.6 & 81.6 & 33.3 & \textbf{28.3} \\

& ESR
& 100.0 & 0.0 & 0.0 & -33.3 & 0.0 & 0.0 & 0.0
& 100.0 & 33.3 & 0.0 & 0.0 & 0.0 & 0.0 & 0.0
& 0.0 & 0.0 & 0.0 & 100.0 & 0.0 & 0.0 & 100.0
& 0.0 & 18.2 \\

& TC-ESR
& 66.7 & 100.0 & 0.0 & 0.0 & 0.0 & 0.0 & 0.0 & 66.7 & 100.0 & 0.0 & 33.3 & 0.0 & 0.0 & 0.0 & 0.0 & -100.0 & 0.0 & 100.0 & 0.0 & 100.0 & 0.0 & 33.3 & 22.7 \\

\bottomrule
\end{tabular}%
}
}
\end{table*}

Table~\ref{tab:main_table_full} reports the mean results over the 22 target tasks.
In addition to the SR and ORNR results presented in the main paper, we report $R_5$ and $\tau_5$, with TC-ESR achieving the lowest SR and ORNR and SWAB achieving the highest $R_5$ and $\tau_5$.
The complete per-task results for all four metrics are reported in Table~\ref{tab:per_task_results}.

\subsection{Task Similarity Analysis}
\begin{figure*}[!t]
\centering
\includegraphics[width=\textwidth]{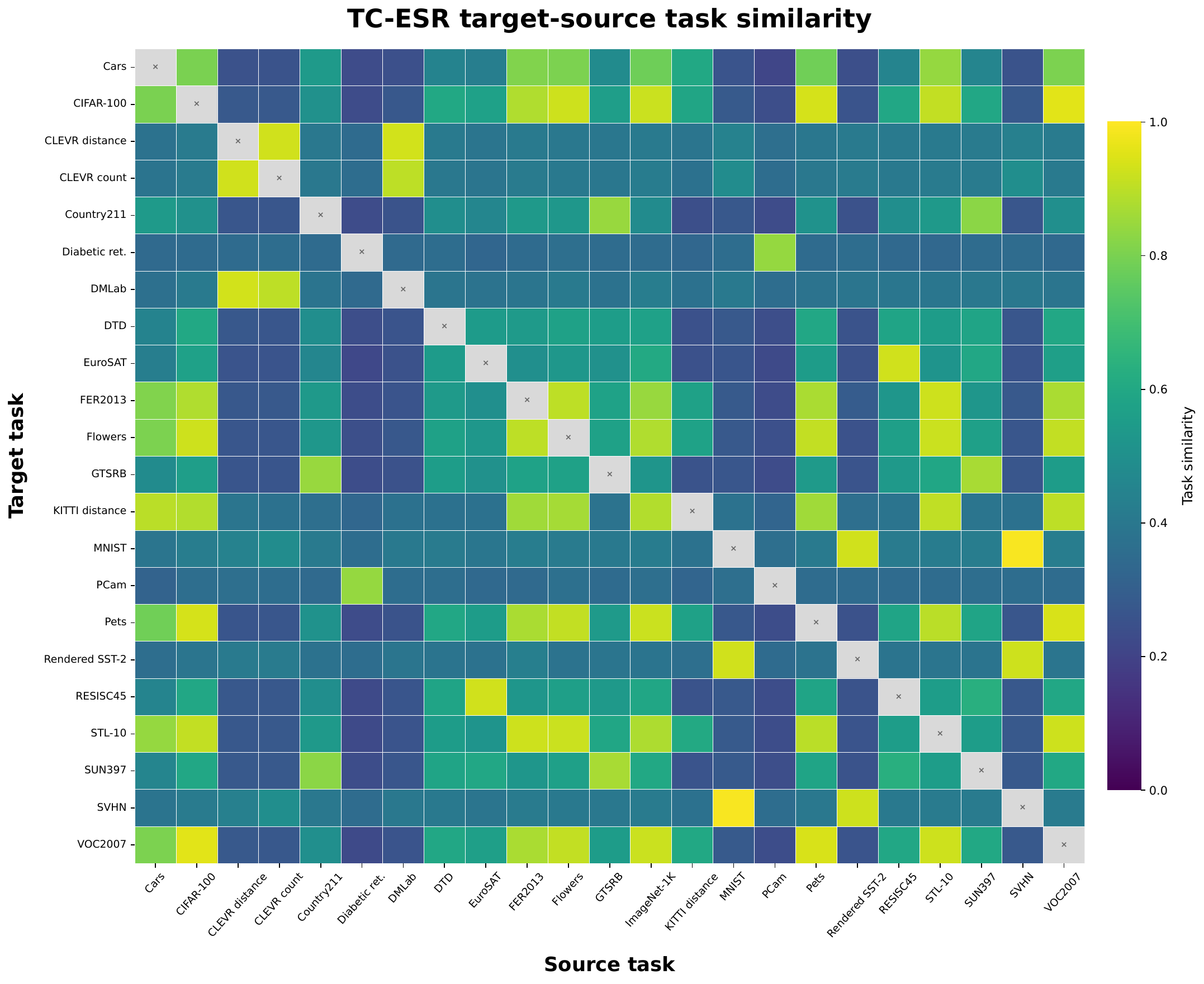} 
\caption{Task similarity matrix used by TC-ESR.
Rows denote target tasks and columns denote source tasks, and each cell reports $S_t(s)$.
The score combines visual domain, label semantics, and label granularity when applicable.
Crosses mark excluded pairs with $s=t$, while ImageNet-1K appears only as a source task under the revised evaluation protocol.
Because the inclusion of label granularity depends on the label scale type of the target task, the matrix need not be symmetric.}
\label{fig:task_sim_heatmap}
\end{figure*}

Figure~\ref{fig:task_sim_heatmap} visualizes the task similarity scores used by TC-ESR.
High similarity is observed between the two CLEVR tasks and between EuroSAT and RESISC45, consistent with their shared synthetic scene and remote sensing domains, respectively.
MNIST and SVHN are also highly similar despite their different visual appearances, reflecting their common digit label semantics.
For Cars, the high similarity to Pets and Flowers is consistent with the combined effects of their shared domain family and similar label granularity, despite weak label semantic overlap.
Together, these patterns indicate that the measured task similarity $S_t(s)$ captures complementary information from visual domain, label semantics, and label granularity.

\subsection{Robustness of ORNR}

\begin{table}[t]
\centering
\caption{Mean normalized regret over 22 targets under alternative denominators. ORNR corresponds to
1.5\,IQR. Lower is better; bold and underline denote the best and second-best results, respectively.}
\label{tab:ornr_denominator_robustness}
\small
\begin{tabular}{llcccccc}
\toprule
Type & Denominator & INB & AvgRank & ModelGPT & SWAB & ESR & TC-ESR \\
\midrule
One-sided & Min--max                  & 21.67 & 22.26 & 21.35 & 22.16 & \underline{19.29} & \textbf{15.97} \\
          & $P_5$                     & 26.99 & 28.65 & 26.61 & 28.78 & \underline{23.78} & \textbf{19.52} \\
          & $P_{10}$                  & 29.35 & 31.19 & 28.97 & 31.26 & \underline{26.58} & \textbf{21.47} \\
          & 1.0\,IQR                  & 26.05 & 28.48 & 25.78 & 28.07 & \underline{23.63} & \textbf{19.01} \\
          & 1.5\,IQR (ORNR)           & 24.66 & 25.69 & 24.25 & 25.81 & \underline{22.56} & \textbf{17.81} \\
          & 2.0\,IQR                  & 24.20 & 25.19 & 23.79 & 25.23 & \underline{21.82} & \textbf{17.41} \\
          & 3.0\,IQR                  & 23.53 & 24.81 & 23.12 & 24.55 & \underline{21.21} & \textbf{16.83} \\
          & 4.0\,IQR                  & 22.67 & 24.81 & 22.51 & 23.70 & \underline{21.11} & \textbf{16.79} \\
          & 5.0\,IQR                  & 22.38 & 23.91 & 22.16 & 23.41 & \underline{20.70} & \textbf{16.50} \\
\midrule
Two-sided & $P_{95} - P_5$            & 30.53 & 33.17 & 30.10 & 32.25 & \underline{27.95} & \textbf{22.56} \\
          & $P_{90} - P_{10}$         & 37.68 & 40.33 & 37.17 & 39.53 & \underline{35.42} & \textbf{28.66} \\
          & Tukey, $k = 1.5$          & 25.02 & 25.87 & 24.59 & 26.06 & \underline{22.94} & \textbf{18.18} \\
          & Tukey, $k = 3.0$          & 23.53 & 24.81 & 23.12 & 24.55 & \underline{21.21} & \textbf{16.83} \\
\bottomrule
\end{tabular}
\end{table}

\begin{table}[t]
\centering
\caption{Mean absolute change in normalized regret when weak candidates are perturbed.}
\label{tab:ornr_perturbation}
\small
\begin{tabular}{lccc}
\toprule
Perturbation & Min--max & ORNR & Reduction (\%) \\
\midrule
Weakest candidate $\rightarrow$ 50\% & 4.58 & 1.19 & 74.0 \\
Weakest candidate $\rightarrow$ 0    & 6.54 & 1.17 & 82.0 \\
Bottom 5\% removed                   & 3.48 & 1.82 & 47.7 \\
Bottom 10\% removed                  & 5.36 & 3.62 & 32.5 \\
Bottom 20\% removed                  & 7.84 & 6.00 & 23.5 \\
\bottomrule
\end{tabular}
\end{table}

\label{app:robust_ornr}
\paragraph{Alternative Denominators.}
Table~\ref{tab:ornr_denominator_robustness} examines whether the comparison depends on the denominator
used by ORNR.
We consider one-sided variants, which replace only the lower end of the range, and two-sided variants,
which also replace its upper end.
For the one-sided variants, in addition to the full range, we consider percentile and interquartile
range based alternatives.
For the percentile alternative $P_q$, we use the $q$th percentile of the candidate model accuracies,
which defines the cutoff for the lowest $q$\% of accuracies, as the lower endpoint.
For the interquartile range alternatives, we define $\mathrm{IQR}_t=Q_{3,t}-Q_{1,t}$ and, for a
multiplier $c$, set the cutoff to $F^{-}_{t,c}=Q_{1,t}-c\,\mathrm{IQR}_t$.
We then use the lowest candidate model accuracy that is no smaller than $F_{t,c}^{-}$ as the lower
endpoint of the denominator.
The multipliers $1.5$ and $3.0$ are commonly used in exploratory data analysis, particularly in
boxplots, to identify outliers and extreme outliers, respectively~\citep{tukey1977eda}.
We additionally vary $c$ from $1.0$ to $5.0$ to evaluate a broader range of choices.
For the two-sided variants, we use the difference between the upper and lower percentiles,
$P_{95}-P_5$ and $P_{90}-P_{10}$, or the range between the upper and lower Tukey whiskers with
multiplier $k$.
TC-ESR achieves the lowest normalized regret under all 13 denominator choices, with ESR consistently
ranking second.
The two-sided Tukey range with $k=3.0$ yields the same values as the one-sided $3.0\,\mathrm{IQR}$
variant, indicating that no candidate is flagged as a high outlier at this threshold.
These results show that the relative advantage of TC-ESR does not depend on the denominator choice.
We adopt $1.5\,\mathrm{IQR}$ as the primary setting because it is the most commonly used criterion,
rather than selecting the multiplier based on these results.

\paragraph{Sensitivity to Weak Candidates.}
This experiment empirically examines the scenario illustrated in Figure~\ref{fig:ornr}b, in which a
single unusually weak candidate inflates the normalization range.
Since neither the best model nor the selected model changes in this scenario, an ideal normalized regret
should remain unchanged.
We therefore modify only the weak candidates of each target, either by lowering the accuracy of the
weakest candidate to half of its value or to zero, or by removing the bottom 5\%, 10\%, or 20\% of
candidates, and measure the mean absolute change in the normalized regret of the selected models under
min--max normalization and ORNR.
Table~\ref{tab:ornr_perturbation} reports these changes together with the relative reduction achieved
by ORNR.
When only the weakest candidate is degraded, ORNR changes 74.0\% to 82.0\% less than min--max
normalization and remains unchanged on 16 of the 22 targets.
When multiple weak candidates are removed, ORNR remains less sensitive, but the reduction decreases from
47.7\% to 23.5\%, because removing many candidates shifts the quartiles that determine its lower whisker.
Thus, ORNR is robust to individual weak outliers but, like any task-relative normalization, is not
invariant to substantial changes in the composition of the candidate pool.

\subsection{Baselines Retuned for Selection Quality}
\label{app:baseline_retune}
In the main experiments, ModelGPT and SWAB select their input feature configurations by
$R_5 + \tau_5$ on source tasks, following their original protocols, whereas TC-ESR selects its
temperature by ORNR.
To rule out that this asymmetry favors TC-ESR, we rerun the same source-only feature selection for
ModelGPT and SWAB with SR or ORNR as the validation objective.
Table~\ref{tab:baseline_retune} shows that this barely changes ModelGPT and degrades SWAB on both SR
and ORNR.
TC-ESR and ESR therefore remain the best and second-best methods regardless of the objective used to
tune the baselines.

\begin{table}[t]
\centering
\caption{ModelGPT and SWAB with feature selection retuned by SR or ORNR.}
\label{tab:baseline_retune}
\small
\begin{tabular}{llcccc}
\toprule
Method & Feature selection & SR & ORNR & $R_5$ & $\tau_5$ \\
\midrule
ModelGPT & $R_5 + \tau_5$ & 7.11 & 24.25 & 41.1 & 18.0 \\
         & SR             & 7.14 & 24.16 & 42.1 & 17.1 \\
         & ORNR           & 7.12 & 24.18 & 42.5 & 16.6 \\
\midrule
SWAB     & $R_5 + \tau_5$ & 7.39 & 25.81 & 47.4 & 28.3 \\
         & SR             & 8.58 & 28.40 & 34.3 & 10.2 \\
         & ORNR           & 8.10 & 26.42 & 36.6 & 10.9 \\
\midrule
ESR      & --             & 5.83 & 22.56 & 31.8 & 18.2 \\
TC-ESR   & --             & 4.66 & 17.81 & 40.0 & 22.7 \\
\bottomrule
\end{tabular}
\end{table}

\section{Role of Target-Conditioned Weighting}
\label{app:discussion}

Prior methods and ESR select nearly the same candidate model for all target tasks. 
This section examines how target-conditioned weighting allows TC-ESR to depart from this choice and how this affects its gains.

\paragraph{The best fixed model as a reference.}

\begin{table}[t]
\centering
\small
\caption{Distribution of the selected candidate models over 22 target tasks. For ModelGPT and SWAB, selections are pooled over ten seeds.}
\label{tab:selection_dist}
\begin{tabular}{lcc}
\toprule
Method & Distinct candidate models & Share of the most frequent (\%) \\
\midrule
INB      & 1 & 100.0 \\
AvgRank  & 1 & 100.0 \\
ModelGPT & 3 & 95.0 \\
SWAB     & 5 & 80.9 \\
ESR      & 2 & 95.5 \\
TC-ESR   & 8 & 45.5 \\
\bottomrule
\end{tabular}
\end{table}

The best fixed model selects the same candidate model for all target tasks using their target accuracies.
It therefore attains the lowest mean SR among all selections that do not depend on the target task. 
As shown in Table~\ref{tab:selection_dist}, prior methods and ESR are close to such selections: INB and AvgRank select the same candidate model for all 22 targets, ModelGPT and SWAB select their most frequent candidate model in 95.0\% and 80.9\% of their selections over targets and seeds, and ESR selects the best fixed model for 21 of the 22 targets using source tasks alone, which largely explains its gain over prior methods. 
In contrast, TC-ESR selects eight distinct candidate models by assigning target-conditioned weights to source tasks. 
Such target-specific selection is necessary to attain lower SR than the best fixed model.

\paragraph{Case study: Retinopathy.}

\begin{table}[t]
\centering
\small
\caption{SR and ORNR on Retinopathy with and without PCam in the source tasks.}
\label{tab:pcam_removal}
\begin{tabular}{lcccc}
\toprule
& \multicolumn{2}{c}{SR (\%p) $\downarrow$} & \multicolumn{2}{c}{ORNR (\%) $\downarrow$} \\
\cmidrule(lr){2-3} \cmidrule(lr){4-5}
Method & w/ PCam & w/o PCam & w/ PCam & w/o PCam \\
\midrule
INB      & 49.81 & 49.81 & 70.37 & 70.37 \\
AvgRank  & 42.93 & 42.93 & 60.65 & 60.65 \\
ModelGPT & 49.81 & 49.81 & 70.37 & 70.37 \\
SWAB     & 49.13 & 49.13 & 69.40 & 69.40 \\
ESR      & \underline{20.57} & \textbf{20.57} & \underline{29.06} & \textbf{29.06} \\
TC-ESR   & \textbf{2.22}  & \textbf{20.57} & \textbf{3.13}  & \textbf{29.06} \\
\midrule
Best fixed model & 20.57 & 20.57 & 29.06 & 29.06 \\
\bottomrule
\end{tabular}
\end{table}

On Retinopathy, PCam is the only source task sharing its domain family, medical images. Its similarity to the target therefore exceeds those of the other source tasks by a large margin, and it receives more than 99.99\% of the source weight. TC-ESR then selects OpenAI RN50x16. As shown in Table~\ref{tab:pcam_removal}, its SR is 2.22\%p, whereas the best fixed model, which ESR also selects, incurs a substantially higher SR of 20.57\%p. Removing PCam from the source and validation tasks and repeating the source-only validation increases the SR of TC-ESR to 20.57\%p, while the selections of the other methods remain unchanged. Among the compared methods, only TC-ESR measures domain similarity between tasks, and thus only it identifies PCam as the source task most relevant to Retinopathy, which accounts for its low SR on this target. 

\paragraph{Concentration of source weights.}

\begin{table}[t]
\centering
\caption{Effective number of source tasks $N_{\mathrm{eff},t}$ of TC-ESR for each target, grouped by domain family. \#Sources denotes the number of source tasks sharing the target's domain family, counting ImageNet-1K. For every target with at least one such source, these sources receive more than 99.3\% of the total weight.}
\label{tab:neff}
\small
\begin{tabular}{llcc}
\toprule
Domain family & Target & \#Sources & $N_{\mathrm{eff},t}$ \\
\midrule
Object centered natural photographs & Stanford Cars          & 8 & 4.24 \\
                                    & CIFAR100               & 8 & 3.93 \\
                                    & FER2013                & 8 & 3.63 \\
                                    & Flowers102             & 8 & 5.25 \\
                                    & KITTI closest distance & 8 & 6.24 \\
                                    & Oxford Pets            & 8 & 4.27 \\
                                    & STL10                  & 8 & 5.22 \\
                                    & VOC2007                & 8 & 3.91 \\
\midrule
Scene centered natural photographs  & Country211             & 2 & 1.80 \\
                                    & GTSRB                  & 2 & 1.77 \\
                                    & SUN397                 & 2 & 1.44 \\
\midrule
Synthetic or rendered images        & CLEVR closest distance & 2 & 2.00 \\
                                    & CLEVR count all        & 2 & 1.65 \\
                                    & DMLab                  & 2 & 1.61 \\
\midrule
Document or glyph images            & MNIST                  & 2 & 1.24 \\
                                    & Rendered SST2          & 2 & 1.99 \\
                                    & SVHN                   & 2 & 1.21 \\
\midrule
Aerial or satellite images          & EuroSAT                & 1 & 1.01 \\
                                    & RESISC45               & 1 & 1.00 \\
\midrule
Medical images                      & Diabetic Retinopathy   & 1 & 1.00 \\
                                    & PCam                   & 1 & 1.00 \\
\midrule
Texture centered photographs        & DTD                    & 0 & 7.97 \\
\bottomrule
\end{tabular}
\end{table}

To examine how concentrated the source weights of TC-ESR are, we compute the effective number of sources for each target,
\begin{equation}
N_{\mathrm{eff},t} = \frac{1}{\sum_{s \in \mathcal{S}_t} w_t(s; \lambda_t^\star)^2},
\end{equation}
where $w_t(s; \lambda_t^\star)$ denotes the source weights in Equation~\ref{eq:tcesr_weight}. This quantity is the effective sample size of \citet{kish1965survey}: it equals one when a single source task receives all of the weight and $|\mathcal{S}_t|$ when the weights are uniform, as in ESR, and can be read as the number of equally weighted source tasks that would yield the same concentration. As shown in Table~\ref{tab:neff}, source tasks sharing the target's domain family receive nearly all of the weight, and the weight is further differentiated among them by label semantics and, where applicable, label granularity. $N_{\mathrm{eff},t}$ is therefore at most 1.01 for the four targets with a single such source, including Retinopathy, whereas it is at least 1.21 for all other targets and ranges from 3.63 to 6.24 for the object-centered targets. This behavior differs from the fixed Nearest-source rule, which performs worse than ESR in Table~\ref{tab:simple_baselines}: TC-ESR concentrates on a single source task only when the target shares its domain family with a single source task.

\paragraph{Results excluding Retinopathy.}

\begin{table}[t]
\centering
\small
\caption{Mean SR and ORNR over 21 target tasks excluding Retinopathy. Bold and underline denote the best and second-best LOVM methods.}
\label{tab:excl_retino}
\begin{tabular}{lcc}
\toprule
Method & SR (\%p) $\downarrow$ & ORNR (\%) $\downarrow$ \\
\midrule
INB      & 5.20 & 22.48 \\
AvgRank  & 5.65 & 24.03 \\
ModelGPT & \underline{5.08} & \underline{22.06} \\
SWAB     & 5.40 & 23.73 \\
ESR      & 5.13 & 22.25 \\
TC-ESR   & \textbf{4.78} & \textbf{18.51} \\
\midrule
Best fixed model & 4.81 & 21.19 \\
\bottomrule
\end{tabular}
\end{table}

We next examine whether the advantage of TC-ESR extends beyond Retinopathy.
Table~\ref{tab:excl_retino} reports the results over the remaining 21 targets, where we exclude Retinopathy only from the evaluation targets and keep the predictions of all methods and the selected $\lambda^\star_t$ of TC-ESR unchanged.

The best fixed model is reselected over the 21 targets, and selecting it by either SR or ORNR yields the same candidate model as before. 
TC-ESR remains the best LOVM method on both metrics, but its SR margin becomes small: its SR of 4.78\%p compares with 5.08\%p for ModelGPT, the strongest prior method on these targets, and 5.13\%p for ESR, and is comparable to the 4.81\%p of the best fixed model, which requires target accuracies. The margin remains substantial in ORNR, where TC-ESR attains 18.51\%, compared with 22.06\% for ModelGPT, 22.25\% for ESR, and 21.19\% for the best fixed model.

\paragraph{Summary.}
While prior methods and ESR select nearly the same candidate model for all target tasks, TC-ESR selects target-specific models by concentrating the source weights on tasks sharing the target's domain family and differentiating among them by label similarity.
Unlike the other methods, TC-ESR identifies the closely related source task of Retinopathy through domain similarity, attaining an SR of 2.22\%p compared with 20.57\%p for the best fixed model.
On the remaining targets, it still attains the lowest mean SR and ORNR among LOVM methods, and matches the best fixed model in SR while surpassing it in ORNR, although the best fixed model requires target accuracies.
These results are consistent with the premise of TC-ESR that models performing well on source tasks similar to the target also tend to perform well on the target.
\end{document}

%% file: math_commands.tex
\usepackage{amsmath,amsfonts,bm}

\def\eqref#1{equation~\ref{#1}}

\def\1{\bm{1}}

\DeclareMathAlphabet{\mathsfit}{\encodingdefault}{\sfdefault}{m}{sl}
\SetMathAlphabet{\mathsfit}{bold}{\encodingdefault}{\sfdefault}{bx}{n}